\documentclass{article}

\usepackage[preprint]{neurips_2025}

\usepackage[most]{tcolorbox}
\usepackage{caption}
\usepackage{geometry}
\usepackage[utf8]{inputenc} 
\usepackage[T1]{fontenc}    
\usepackage{hyperref}       
\usepackage{url}            
\usepackage{booktabs}       
\usepackage{amsfonts}       
\usepackage{nicefrac}       
\usepackage{microtype}      
\usepackage{xcolor}         
\usepackage{amsmath}
\usepackage{arydshln}
\usepackage{amssymb}
\usepackage{textcomp}
\usepackage{natbib}
\usepackage{enumitem}
\usepackage{xspace}
\usepackage[table]{xcolor}
\usepackage{algorithm}
\usepackage{algpseudocode}
\usepackage{CJKutf8} 

\definecolor{PromptGreen}{rgb}{0.00,0.55,0.25}
\newcommand{\varnumq}{\texttt{\{num\_q\}}}
\newcommand{\varparagraph}{\texttt{\{paragraph\}}}
\newcommand{\varquestion}{\texttt{\{current\_question\}}}
\definecolor{bonus_green}{RGB}{0,100,0}
\newcommand{\bbonus}[1]{{\textcolor{bonus_green}{$^{\uparrow#1}$}}}

\newcommand{\drer}{DRER\xspace}
\newcommand{\logictree}{LogicTree\xspace}

\newcommand{\fbeta}{\emph{F}\(\beta\)\emph{-Score}\xspace}

\title{Rethinking Reasoning Quality in Large Language Models through Enhanced  Chain-of-Thought via RL}

\author{%
  Haoyang He \quad Zihua Rong \quad Kun Ji \quad Chenyang Li \quad Qing Huang \\
  \textbf{Chong Xia} \quad \textbf{Lan Yang} \quad \textbf{Honggang Zhang}\thanks{Correspondence to zhhg@bupt.edu.cn} \\
  Beijing University of Posts and Telecommunications\\
}

\begin{document}
\begin{CJK}{UTF8}{gkai}

\maketitle

\begin{abstract}
Reinforcement learning (RL) has recently become the dominant paradigm for strengthening the reasoning abilities of large language models (LLMs). Yet the rule-based reward functions commonly used on mathematical or programming benchmarks assess only answer format and correctness, providing no signal as to whether the induced Chain-of-Thought (CoT) actually improves the answer. Furthermore, such task-specific training offers limited control over logical depth and therefore may fail to reveal a model’s genuine reasoning capacity. We propose \textbf{D}ynamic \textbf{R}easoning \textbf{E}fficiency \textbf{R}eward (\textbf{\drer}) — a plug-and-play RL reward framework that reshapes both reward and advantage signals. (i) A \emph{Reasoning Quality Reward} assigns fine-grained credit to those reasoning chains that demonstrably raise the likelihood of the correct answer, directly incentivising the trajectories with beneficial CoT tokens. (ii) A \emph{Dynamic Length Advantage} decays the advantage of responses whose length deviates from a validation-derived threshold, stabilising training. To facilitate rigorous assessment, we also release \emph{\logictree}, a dynamically constructed deductive reasoning dataset that functions both as RL training data and as a comprehensive benchmark. Experiments confirm the effectiveness of \drer: our 7B model attains GPT-o3-mini level performance on \logictree with 400 trianing steps, while the average confidence of CoT-augmented answers rises by 30\%. The model further exhibits generalisation across diverse logical-reasoning datasets, and the mathematical benchmark AIME24. These results illuminate how RL shapes CoT behaviour and chart a practical path toward enhancing formal-reasoning skills in large language models. All code and data are available in repository \url{https://github.com/Henryhe09/DRER}.
\end{abstract}

\section{Introduction}
Recent reasoning models~\cite{gemini-thinking,qwq,k1.5}, including R1-like reproductions~\cite{k1.5,areal,dapo,shao2024deepseekmath,hu2025reinforce++,rloo,ahmadian2024basicsrevisitingreinforcestyle,sutton1998rl}, have adopted reinforcement learning (RL) to enhance chain-of-thought reasoning. By systematically exploring verifiable reasoning paths that lead to correct answers, these methods incrementally boost performance and deliver remarkable gains. Current RL-driven CoT approaches typically train on mathematics and programming benchmarks~\cite{o1,guo2025deepseek,cobbe2021training,
chen2021evaluating}, whose inherently stepwise solution procedures serve as natural proxies for logical inference~\cite{wang2024measuring,
li2024chain}, and they rely on rule-based reward~\cite{o1,guo2025deepseek} functions that assess only final answer correctness and formatting. This reliance stems from the straightforward evaluability of math and code tasks, where simple answer extraction or format checks suffice to assign reward signals and compute policy advantages.

However, this approach still faces two critical challenges. First, by relying solely on final‐answer correctness as the reward signal, the model cannot distinguish which reasoning steps statistically boost the likelihood of the correct answers~\cite{paul2024making}, nor quantify each token’s substantive contribution to the conclusion; instead, it may lean on “decorative” chains that diverge from genuine deductive paths~\cite{zhang2024rest}, thereby undermining the accurate evaluation and effective training of its reasoning ability.

Second, the corpora used to reinforce “reasoning ability” are almost entirely drawn from execution‐verifiable domains~\cite{sprague2024cot}—such as mathematical problem sets and code synthesis tasks—while unified training data targeting pure formal logical inference remains severely lacking~\cite{morishita2024enhancing}. Such constrained training regimens risk conceptual overextension~\cite{paul2024making}, whereby success on specific tasks is misconstrued as evidence of broadly applicable logical reasoning skills, potentially leading to an overestimation of the model’s true inferential competence.

\begin{figure}[h]
  \centering
  \includegraphics[width=\linewidth]{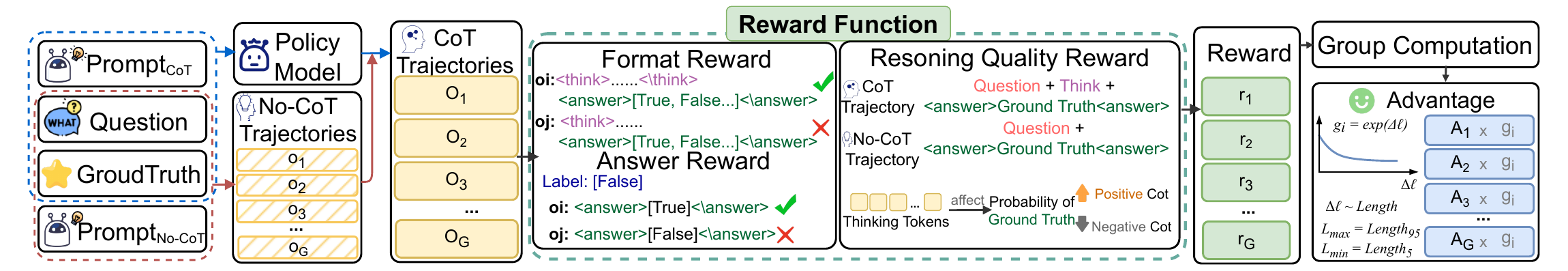}
  \caption{Overview of the Dynamic Reasoning Efficiency Reward (\drer) framework. \(\text{Length}_{95}\) and \(\text{Length}_{5}\) represent the 95th and 5th percentile lengths, respectively, computed from the validation set, and are used to normalize reasoning trajectory lengths according to task type or difficulty.}
  \label{fig:full_algorithm}
\end{figure}

To address the limitations of outcome‐only reward modeling in reasoning tasks, we propose \emph{Dynamic Reasoning Efficiency Reward} (\drer), a plug‐and‐play reinforcement learning framework that reshapes both reward and advantage signals. \drer introduces two key mechanisms: (1) a \emph{Reasoning‐Quality Reward}, which assigns fine‐grained credit to reasoning chains that statistically improve the likelihood of the correct answer, thereby reinforcing the utility of CoT tokens; and (2) a \emph{Dynamic‐Length Advantage}, which attenuates the policy advantage of responses whose lengths deviate from a validation‐derived threshold, improving training stability. The overall framework is illustrated in Figure~\ref{fig:full_algorithm}. In addition, we release \emph{\logictree}, a domain‐agnostic deductive reasoning dataset carefully constructed to provide focused training supervision and to serve as a clean evaluation benchmark for identifying pathological reasoning behaviours.

Our experiments show that DRER delivers a marked leap in deductive skill: Qwen-2.5-7B-Instruct-1M model trained on \logictree with 400 steps raises overall accuracy from 13\% to 60\%, still solves 31\% of the hardest depth-8 items, and boosts answer-confidence by roughly 60\% while reducing token consumption by 75\% relative to DAPO/GRPO baselines.

The main contributions of this paper are summarized as follows:
\begin{itemize}[leftmargin=15pt, itemsep=0pt, topsep=0pt, partopsep=0pt, parsep=0.5pt]
  \item We propose \textbf{\drer} (\textbf{D}ynamic \textbf{R}easoning \textbf{E}fficiency \textbf{R}eward), a novel reinforcement learning rewawrd framework that adaptively reshapes both reward and advantage signals to improve CoT reasoning.
  \item We release \emph{\logictree}, a domain-agnostic benchmark for formal deductive reasoning that serves dual purposes: functioning as both a focused training set and a clean evaluation benchmark, while providing highlight insights into LLMs reasoning behaviours.
  \item We systematically validate our approach through extensive experiments, confirming the effectiveness of our methodology in improving both reasoning quality and efficiency.
\end{itemize}

\section{Preliminary}
\paragraph{Modeling Language Generation as a Token-Level MDP}

Reinforcement learning aims to learn a policy that maximizes cumulative reward through interaction with an environment. We model language generation as a sequential decision process within a Markov Decision Process (MDP) framework~\citep{ouyang2022training}. Let \(x = (x_0, \dots, x_m)\) be the input prompt and \(y = (y_0, \dots, y_T)\) the generated response, with both drawn from a finite vocabulary \(\mathcal{A}\). At step \(t\), the state is \(s_t = (x_0, \dots, x_m, y_0, \dots, y_t)\), and the action \(a = y_{t+1} \in \mathcal{A}\) selects the next token. Transitions are deterministic: \(\mathbb{P}(s_{t+1} \mid s_t, a) = 1\), where \(s_{t+1} = (x_0, \dots, x_m, y_0, \dots, y_{t+1})\). Generation ends upon producing a terminal token \(\omega\). The reward function \(R(s, a)\) provides scalar feedback on output quality. The initial state \(s_0\) is the tokenized prompt, sampled from a distribution \(d_0\) over inputs. This MDP formulation allows reinforcement learning—both value-based and value-free—to align language model generation with desired objectives and human preferences.
\paragraph{Proximal Policy Optimization (PPO)}
Proximal Policy Optimization (PPO)~\citep{schulman2017proximalpolicyoptimizationalgorithms} is a widely–used
value–based policy–gradient algorithm that stabilises training by
constraining the magnitude of each update.  Given a mini–batch of prompt-answer
pairs \((x,y^\star)\sim\mathcal{D}\) and roll-outs
\((s_t,a_t)\sim\pi_{\theta_{\text{old}}}\!\left(\cdot\mid x\right)\),
PPO maximises the \emph{clipped surrogate objective}
\begin{equation}
\begin{aligned}
\label{eq:ppo_loss}
\mathcal{J}_{\text{PPO}}(\theta)=
\mathbb{E}_{(q,a)\sim \mathcal{D},o_{\le t}\sim\pi_{\theta_{\text{old}}}(\cdot\mid q)}
\Bigl[
\min\!\bigl(
r_t(\theta)\,\hat{A}_t,\;
\operatorname{clip}\!\bigl(r_t(\theta),1-\varepsilon,1+\varepsilon\bigr)\,\hat{A}_t
\bigr)
\Bigr],
\end{aligned}
\end{equation}
\begin{equation}
\begin{aligned}
r_t(\theta)=\frac{\pi_\theta(a_t\mid s_t)}{\pi_{\theta_{\text{old}}}(a_t\mid s_t)},
\end{aligned}
\end{equation}
where $(q, a)$ denotes a question–answer pair sampled from the data distribution $\mathcal{D}$, \(r_t(\theta)\) is the importance–sampling ratio and
\(\varepsilon\) defines the trust region.
The term \(\hat{A}_t\) is an estimate of the advantage at step \(t\),
computed from the value function \(V_\phi\) and the sampled rewards via
\emph{Generalised Advantage Estimation (GAE)}~\citep{schulman2018highdimensionalcontinuouscontrolusing}:
\begin{equation}
\begin{aligned}
\hat{A}^{\text{GAE}(\gamma,\lambda)}_t
      &=\sum_{l=0}^{\infty}(\gamma\lambda)^l\,\delta_{t+l},
      \label{eq:gae}\\
\end{aligned}
\end{equation}
where
\begin{equation}
    \delta_{l}=R_l+\gamma V(s_{l+1})-V(s_l),\quad 0\le\gamma,\lambda\le1.
\end{equation}
$(q, a)$ denotes a question–answer pair sampled from the data distribution $\mathcal{D}$.Here \(\gamma\) is the discount factor, \(\lambda\) controls the
bias–variance trade-off of GAE, and \(r_t\) denotes the reward obtained
at step \(t\).
\paragraph{Group Relative Policy Optimization (GRPO)}

GRPO\citep{shao2024deepseekmath} removes the value function used in PPO and estimates the advantages within a group of $G$ responses sampled by the behavior policy $\pi_{\theta_\text{old}}$ for each pair of questions-answers $(q,a)$. 
GRPO maximizes a PPO-style clipped objective with an explicit KL penalty:
\begin{equation}
\label{eq:grpoloss}
\begin{aligned}
\mathcal{J}_\text{GRPO}(\theta)
&= \mathbb{E}_{(q,a)\sim\mathcal{D},\, \{o_i\}\sim\pi_{\theta_\text{old}}}
\\
&\Biggl[
\frac{1}{G}\sum_{i=1}^{G}\frac{1}{|o_i|}
\sum_{t=1}^{|o_i|}
\Bigl(
\min\!\bigl(r_{i,t}\hat{A}_{i,t},\,
           \text{clip}(r_{i,t}, 1{-}\epsilon, 1{+}\epsilon)\hat{A}_{i,t}
\bigr)
- \beta\,\mathrm{D}_{\mathrm{KL}}(\pi_\theta \parallel \pi_\text{ref})
\Bigr)
\Biggr],
\end{aligned}
\end{equation}
where
\begin{equation}
    r_{i,t}(\theta)=\frac{\pi_{\theta}(o_{i,t} \mid q, o_{i,<t})}{\pi_{\theta_{\text{old}}}(o_{i,t} \mid q,o_{i,<t})},\quad\hat{A}_{i,t} = \frac{R_i - \text{mean}(\{R_i\}_{i=1}^G)}{\text{std}(\{R_i\}_{i=1}^G)}.
\label{eq:advantage_calculation}
\end{equation}

GRPO first averages token-level losses within each response and then across the group, a sample-level aggregation that can implicitly favor longer responses and thus influence training dynamics~\citep{liu2025understandingr1zeroliketrainingcritical}.

\paragraph{Decouple Clip and Dynamic Sampling Policy Optimization (DAPO)}

DAPO\citep{dapo} shares GRPO’s group-based sampling and advantage normalization, but differs in two key aspects.  
First, it replaces GRPO’s symmetric clipping with asymmetric clipping bounds, allowing for unbalanced exploration and conservative updates.  
Second, it introduces a dynamic sampling constraint that requires both correct and incorrect responses in the sampled group to ensure meaningful advantage shaping.  
The resulting objective is:

\begin{equation}
\label{eq:dapoloss}
\begin{aligned}
\mathcal{J}_\text{DAPO}(\theta)
&=  \mathbb{E}_{(q,a)\sim\mathcal{D},\, \{o_i\}\sim\pi_{\theta_\text{old}}}
\\
&\Biggl[
\frac{1}{\sum_{i=1}^G |o_i|} \sum_{i=1}^G \sum_{t=1}^{|o_i|}
\min\!\Bigl(
r_{i,t} \hat{A}_{i,t},\,
\text{clip}(r_{i,t}, 1{-}\varepsilon_\text{low}, 1{+}\varepsilon_\text{high}) \hat{A}_{i,t}
\Bigr)
\Biggr],
\end{aligned}
\end{equation}

where optimization is applied only if the sampled responses are not all equivalent to the reference answer.  
$r_{i,t}$ and $\hat{A}_{i,t}$ are defined as in Equation~\ref{eq:advantage_calculation}.

\paragraph{Reward Modeling}
In the reinforcement learning (RL) for large language models (LLMs), reward modeling is typically categorized into two main approaches: reward models (RM) and rule-based rewards.
Reward models, including outcome and process reward models (PRMs), learn a function through supervised learning, enabling finer-grained evaluation of intermediate reasoning steps. Methods such as as MATH-SHEPHERD~\citep{wang2024mathshepherdverifyreinforcellms}, ReST-MCTS~\citep{zhang2024restmctsllmselftrainingprocess}, and OmegaPRM~\citep{luo2024improvemathematicalreasoninglanguage} show that PRMs can improve the consistency and generalization of reasoning. However, they also introduce higher annotation costs, potential data bias (e.g. MCTS-generated traces), and reduced reliability in early step evaluation, which can affect training stability~\citep{li202512surveyreasoning}.

Rule-based rewards are widely adopted, where simple criteria such as answer correctness and syntactic validity are used to evaluate model outputs. Representative works~\citep{lyu2025exploringlimitoutcomereward,xie2025logicrlunleashingllmreasoning,li2025limrrlscaling,aggarwal2025l1controllinglongreasoning} like DeepSeek-R1~\citep{guo2025deepseek}  utilize correctness-based signals to construct efficient and interpretable training pipelines. The primary advantages of rule-based rewards are twofold: firstly, they exhibit low implementation cost and, secondly, they are characterised by high transparency. These properties render them well-suited for large-scale RL training. However, their limitations are also evident: these methods only evaluate final outcomes, ignoring the quality of intermediate reasoning steps. As a result, models may learn to "shortcut" reasoning, producing correct answers without coherent or logically valid chains of thought—leading to misalignment between reasoning processes and outputs~\citep{zhang2025r1rewardtrainingmultimodalreward}.

\label{preliminary}





\section{Method}

\subsection{\drer}
Rule-based rewards, such as answer correctness and format validity, minimal signals neglect to consider the reasoning trajectory that culminates in the ultimate response. Consequently, they may permit verbose, irrelevant chains of thought, which compromise reasoning transparency and reliability.

In order to address this limitation, a novel reward framework, Dynamic Reasoning Efficiency Reward (\drer), is introduced. This plug-and-play system has been designed to shape not only the correctness of final outputs, but also the efficiency and utility of intermediate reasoning steps.

Given an input question $x$, the large-language model (LLM) $\pi_{\theta}$ produces an output sequence $y$ autoregressively:
\begin{equation}
  \pi_{\theta}(y\mid x)
  \;=\;
  \prod_{t=1}^{T}
  P_{\pi_{\theta}}\!\bigl(y_t \mid x,\, y_{<t}\bigr),
  \label{eq:autoregressive}
\end{equation}
where the sequence $y = [c, a]$ denotes the model’s output sequence, where the first contiguous segment  
$c = (c_{1},\dots,c_{T_{\!c}})$ comprises the CoT tokens and the second segment  
$a = (a_{1},\dots,a_{T_{\!a}})$ contains the answer tokens.  
The overall sequence length satisfies $T = T_{\!c} + T_{\!a}$.

We believe that if the generated CoT tokens $c$ are positive and coherent
with the correct answer, it should \emph{increase} the model’s confidence in
predicting ground-truth answer token:
\begin{equation}
\ell_{\text{CoT}}
      =\frac{1}{T_a}\sum_{t=1}^{T_a}
        \log\pi_{\theta}\!\bigl(a_t^{\star}\mid x_{CoT},\,c,\,a_{<t}^{\star}\bigr),
\quad
\ell_{\text{NoCoT}}
      =\frac{1}{T_a}\sum_{t=1}^{T_a}
        \log\pi_{\theta}\!\bigl(a_t^{\star}\mid x_{NoCoT},\,a_{<t}^{\star}\bigr),
\label{eq:cot_improves_confidence}
\end{equation}
CoT reasoning tokens that positively contribute to the model's ability to infer the correct answer should satisfy
\begin{equation}
\ell_{\text{CoT}} \;>\; \ell_{\text{NoCoT}}.
\label{eq:cot>nocot}
\end{equation}

where $x_{CoT}$ and $x_{NoCoT}$ denote the CoT and no CoT input question respectively; 
$c=(c_{1},\dots,c_{T_c})$ is the generated CoT of length $T_c$;  
$a^{\star}=(a_{1}^{\star},\dots,a_{T_a}^{\star})$ is the ground-truth
answer consisting of $T_a$ tokens, and $a_{<t}^{\star}$ stands for its prefix
up to position $t{-}1$;  
Finally, $\pi_{\theta}$ is the autoregressive language model policy
parameterised by $\theta$.

To validate this hypothesis, we conducted experiments on the GSM8K benchmark using \textsc{Qwen2.5-7B-Instruct-1M}~\citep{qwen2.5}; Detailed experimental setup and results are provided in the section~\ref{sec:val_log}.

\paragraph{Reasoning Quality Reward}
To make the confidence‐boosting property in \eqref{eq:cot>nocot} learnable,
we define for each training instance $\mathbf{x}$ the log-likelihood margin
\begin{equation}
\Delta(\mathbf{x}) \;=\; \ell_{\text{CoT}} - \ell_{\text{NoCoT}},
\end{equation}
where $\ell_{\text{CoT}}$ and $\ell_{\text{NoCoT}}$ are given in
\eqref{eq:cot_improves_confidence}.  
A positive $\Delta(\mathbf{x})$ indicates that the generated
CoT reasoning tokens enhance the model’s confidence in the correct answer,
whereas a negative value reveals detrimental or spurious reasoning.

To obtain a numerically stable reward, we pass the margin through a smooth,
bounded squashing function
\begin{equation}
R_q \;=\; \tanh\!\bigl(\Delta(\mathbf{x})\bigr),
\end{equation}
yielding the \emph{reasoning-quality reward}.  
The hyperbolic tangent preserves the sign of the margin, caps extreme values.

We incorporate $R_q$ into the overall reinforcement-learning objective by
maximising the expected composite return
\begin{equation}
R \;=\; R_{\text{task}} +\lambda_q R_q,
\end{equation}
where $R_{\text{task}}$ denotes the task-level reward
(e.g., answer correctness) and $\lambda_q>0$ is a weighting coefficient that
balances task success and reasoning quality.
This formulation directly rewards reasoning chains that demonstrably increase
the likelihood of the correct answer while penalising uninformative or
misleading chains, thereby systematically improving the model’s logical
reliability and interpretability.

\paragraph{Dynamic Length Advantage}
After every validation round we record the lengths
$\{L_i\}$ of responses that are both correct and structurally valid
within each difficulty bucket\footnote{%
A bucket may correspond to a task type, question template, or any other
granularity used in specific tasks.}.
The empirical \(5\%\) and \(95\%\) quantiles define a dynamic
lower and upper length bound,
\(
L_{\min}^{(d)}
\)
and
\(
L_{\max}^{(d)}
\),
respectively, for bucket \(d\).
For a training sample \(i\) with effective response length \(\ell_i\),
we introduce a multiplicative attenuation coefficient
\begin{equation}
g_i
=
\exp\!\Bigl(
-\,
\frac{
      \max\{0,\,L_{\min}^{(d)}-\ell_i\,,\ell_i-L_{\max}^{(d)}\}
     }{\tau}
\Bigr),
\qquad
\tau>0,
\label{eq:length_decay}
\end{equation}

where \(L_{\min}^{(d)}\) denotes the 5th-percentile response length observed in the previous validation step for bucket \(d\), while \(L_{\max}^{(d)}\) corresponds to the 95th percentile in the same distribution. The variable \(\ell_i\) represents the effective response length of the current sample \(i\), and \(\tau \in [5,10]\) is a temperature hyperparameter that controls the decay rate of the attenuation function.
\noindent

The attenuation is then applied to the advantage
computed by \textsc{Group Computation},
\(
\hat{A}_i = g_i\,A_i,
\)
so that responses that are excessively short
(\(\ell_i < L_{\min}^{(d)}\))
or verbose
(\(\ell_i > L_{\max}^{(d)}\))
are exponentially down-weighted.
This mechanism penalises pathological length behaviours
while preserving the signal of well-sized, high-quality chains of thought. The complete algorithm procedure of DRER is detailed in Appendix~\ref{appendix:DRER}.

\subsection{Dataset}

Prior research typically classifies model reasoning into four cognitive behaviours~\citep{gandhi2025cognitive}: verification, backtracking, sub-goal setting, and backward chaining. While practical, this taxonomy lacks grounding in formal logical theory. Other approaches conflate logical reasoning with mathematical~\citep{shen2025satori,shao2024deepseekmath} computation or coding~\citep{li2025codei}, or overextend "logical capability" into domains like medicine~\citep{jiang2024reasoning,lucas2024reasoning} or law~\citep{wang2024legalreasoner, yue2024lawllm}, neglecting formal deductive reasoning assessment in LLMs.

We address this gap by focusing exclusively on pure deductive reasoning, selecting seven fundamental inference rules from mathematical logic~\ref{appendix:paradigms}. Importantly, \logictree evaluation integrates previously identified cognitive behaviours under rigorous logical constraints rather than excluding them. For instance, proof by contradiction (e.g., given $p\!\rightarrow\!q$ and $\neg q$, infer $\neg p$) requires the model to reflect on prior statements, identify inconsistencies, and draw logical conclusions demonstrating how formal logic frameworks can systematically assess these cognitive processes.

\paragraph{\logictree}

\begin{figure}[h]
  \centering
  \includegraphics[width=\linewidth]{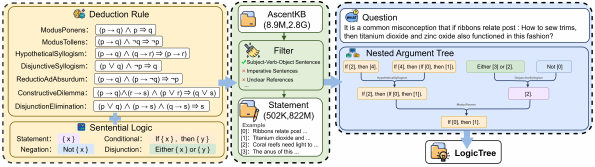}
  \caption{The framework of \textsc{\logictree} automatic construction pipeline. We first sample atomic logic structures and sentences from seven deduction logic rules and four sentential logics, then fill it with natural statements in filtered AscentKB~\citep{nguyen2021advanced}, and eventually construct the nested argument tree. Those intermediate will be hidden and transformed into questions.}
  \label{fig:tree}
\end{figure}

Inspired by LogiQA\citep{logiqa}, PrOntoQA\citep{propontoqa}, PrOntoQA-OOD\citep{prompontoqaood}, JustLogic\citep{justlogic}, our \logictree dataset is constructed on seven inference rules and four sentential logics from mathematical logic, forming a nested binary reasoning tree with dynamically controllable difficulty. It features semantic-logic independence and avoids pre-training data leakage, focusing exclusively on pure deductive reasoning assessment. We significantly enhanced both the construction and evaluation systems beyond those datasets, creating a more reliable, challenging benchmark that provides comprehensive model evaluation.

\logictree incorporates several key improvements: (1) extracting intermediate conclusions as extension questions to assess reasoning completeness and explainability, substantially increasing benchmark difficulty; (2) refining the nested reasoning structure construction process for precise control over depth, width, and question quantity; (3) introducing logical consistency metrics by evaluating models across five questions with identical logical paradigms but different natural language statements; and (4) implementing an \fbeta that integrates answer rate and precision, excluding invalid and \emph{Unknown} responses to reflect accuracy independent of random guessing.

\paragraph{Evaluation}

The \logictree dataset is programmatically generated with parameters controlling scale and difficulty through a mapping table that governs tree width and sub-problem distribution. Unlike traditional NLI datasets with three-class classification~\citep{cheng2025empowering,hanmeng-yue-2024-da} (\emph{True}, \emph{False}, or \emph{Uncertain}), we restrict labels to \emph{True}/\emph{False} to mitigate semantic ambiguity that often artificially inflates accuracy by encouraging defaulting to \emph{Uncertain}. LLMs may respond with \emph{Unknown} during inference, reducing statistical noise from random guessing.

Therefore, we introduce three evaluation metrics: (1) \emph{Accuracy}: Standard correctness rate; (2) \emph{Consistency Ratio}: Stability across logically equivalent queries; (3) \fbeta: Balances Answer Rate (proportion of valid \emph{True}/\emph{False} responses) and Precision (accuracy among valid responses), prioritizing precision while enforcing minimum Answer Rate (>0.3) to discourage overreliance on \emph{Unknown} responses.

\section{Experiment}
\subsection{Experimental Settings}
In our experiments, we use the \logictree dataset, which contains 9,600 questions spanning 8 levels of reasoning depth. The dataset is split into training, validation, and test sets with a ratio of 10:1:1. The most challenging problems involve a reasoning depth of 8 and a branching width of 6, comprising 7 sub-questions derived from intermediate conclusions.To evaluate logical consistency, five questions were constructed for each logical structure by varying the natural language phrasing , ensuring that the model cannot achieve high consistency scores through random guessing.

We post-train Qwen2.5-7B-Instruct-1M using the \logictree dataset within the proposed \drer framework on DAPO for 400 steps. Two baseline algorithms, GRPO and DAPO, are adopted for comparison. More detailed parameters settings can be found in Appendix~\ref{appendix:training_setting}
The task reward (\( R_{\text{task}} \)) consists of two components: a format score and an answer score.

\textbf{Format Score:} The format score (\( S_{\text{format}} \)) evaluates whether the model's response adheres to the required output structure:

\[
S_{\text{format}} =
\begin{cases}
1,  & \text{if format is correct} \\
-1, & \text{if format is incorrect}
\end{cases}
\]

\textbf{Answer Score:} The answer score (\( S_{\text{answer}} \)) evaluates the correctness of the response content against the ground truth:

\[
S_{\text{answer}} =
\begin{cases}
2,   & \text{if the answer fully matches the ground truth} \\
-1.5,& \text{if the answer partially mismatches the ground truth} \\
-2,  & \text{if the answer cannot be parsed or is missing}
\end{cases}
\]

The total task reward is computed as:
\[
R_{\text{task}} = S_{\text{format}} + S_{\text{answer}}
\]

The complete training parameters are detailed in the appendix ~\ref{appendix:training_setting}.

\subsection{Main Results}
\begin{figure}[h]
  \includegraphics[width=\linewidth]{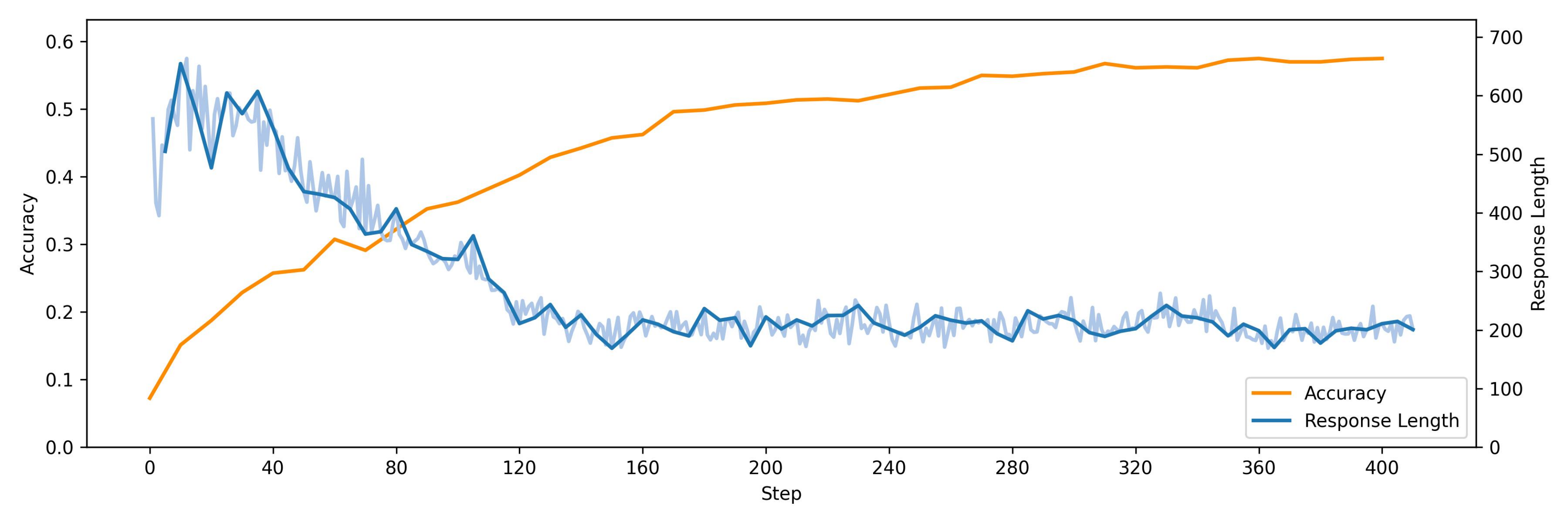}
  \caption{Training dynamics of the DAPO baseline with the \drer framework over 400 steps.}
  \label{fig:train_result}
\end{figure}

\begin{figure}[h]
  \includegraphics[width=\linewidth]{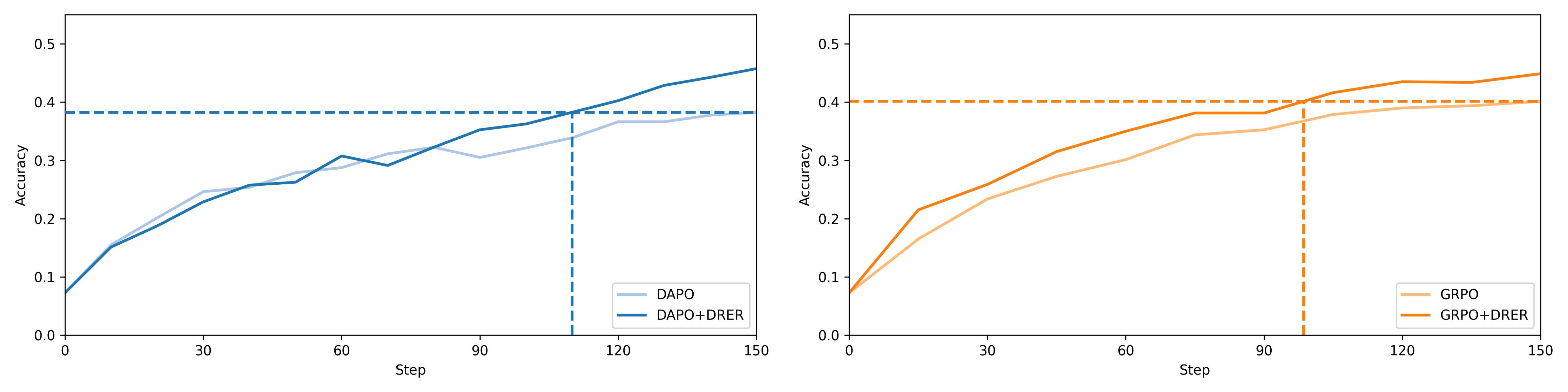}
  \caption{Accuracy on the \logictree during post-training with DAPO (left) and GRPO (right), with and without \drer.}
  \label{fig:acc_compare}
\end{figure}

\paragraph{Training}
Throughout the 400 optimisation steps, we observe a monotonic rise in the model’s accuracy on the \logictree from  \(7\%\) at the outset to nearly \(60\%\) in figure~\ref{fig:train_result}. Detailed evaluation data are in Table~\ref{tab:acc}.In both settings, \drer consistently improves final accuracy and accelerates convergence. Figure~\ref{fig:acc_compare} indicates the step at which the baseline (DAPO or GRPO) reaches its final precision, showing that DRER achieves a significantly higher or comparable performance earlier, highlighting its efficiency in guiding learning through structured reasoning signals.

\paragraph{Evaluation}

\begin{table}[t]

  \centering
  \small
  \caption{Comparsion of Accuracy on \logictree across various logical depth.}
  \label{tab:logic}
  \begin{tabular}{@{}l *{9}{c} @{}}
    \toprule
    Model & 1 & 2 & 3 & 4 & 5 & 6 & 7 & 8 & Avg. \\
    \midrule
    Qwen3-235B-A22B & \textbf{0.96} & \textbf{0.83} & 0.66 & \textbf{0.71} & 0.46 & 0.32 & 0.25 & 0.07 & 0.53 \\
    Deepseek-R1 & 0.85 & 0.76 & 0.61 & 0.47 & 0.36 & 0.18 & 0.19 & 0.07 & 0.44 \\
    Claude-3.7-Sonnet & 0.76 & 0.67 & 0.21 & 0.10 & 0.07 & 0.02 & 0.02 & 0.00 & 0.23 \\
    Qwen3-8B & 0.86 & 0.83 & 0.49 & 0.44 & 0.32 & 0.11 & 0.14 & 0.08 & 0.41 \\
    GPT-o4-mini & 0.74 & 0.64 & 0.25 & 0.20 & 0.10 & 0.06 & 0.05 & 0.02 & 0.26 \\
    GPT-o3-mini & 0.66 & 0.56 & 0.07 & 0.07 & 0.03 & 0.02 & 0.01 & 0.00 & 0.18 \\
    Qwen3-4B & 0.74 & 0.74 & 0.39 & 0.29 & 0.29 & 0.06 & 0.09 & 0.04 & 0.33 \\
    \midrule
    Gemini-2.5-Flash-Preview & 0.86 & 0.64 & 0.41 & 0.31 & 0.24 & 0.11 & 0.06 & 0.00 & 0.33 \\
    GPT-4o & 0.63 & 0.60 & 0.28 & 0.13 & 0.13 & 0.00 & 0.00 & 0.00 & 0.22 \\
    Phi-4-14B & 0.72 & 0.67 & 0.31 & 0.27 & 0.19 & 0.04 & 0.01 & 0.01 & 0.28 \\
    Gemma-3-27B & 0.65 & 0.41 & 0.15 & 0.04 & 0.00 & 0.00 & 0.00 & 0.00 & 0.16 \\
    Deepseek-v3 & 0.39 & 0.24 & 0.05 & 0.06 & 0.00 & 0.00 & 0.00 & 0.00 & 0.09 \\
    GPT-4o-mini & 0.44 & 0.24 & 0.27 & 0.11 & 0.12 & 0.02 & 0.02 & 0.01 & 0.15 \\
    \midrule
    Qwen2.5-7B-Instruct-1M & 0.36 & 0.29 & 0.15 & 0.12 & 0.08 & 0.01 & 0.01 & 0.00 & 0.13 \\
    \rowcolor{cyan!20}
    \textbf{Ours (DAPO+DRER)} & 0.90 & \textbf{0.83} & \textbf{0.76} & 0.59 & \textbf{0.67} & \textbf{0.45} & \textbf{0.31} & \textbf{0.31} & \textbf{0.60}\bbonus{0.47} \\
    \bottomrule
  \end{tabular}
  \label{tab:acc}
\end{table}

As demonstrated Table~\ref{tab:acc}, even advanced models such as GPT-o3-mini, deepseek-r1, and Claude3.7 achieve accuracy scores below \(20\%\) on problems with reasoning depths of 7-8 in the \logictree test set. The best performing model, Qwen3-235B, maintains the highest accuracy of \(25\%\) on problems with reasoning depth of 7, with an average accuracy of \(53\%\). This reveals significant deficiencies in the complex deductive reasoning capabilities of existing reasoning models. In contrast, our trained 7B model achieves state-of-the-art performance in terms of average accuracy, showing substantial improvement over the base model, and maintains a \(31\%\) accuracy rate even at maximum reasoning depth.

In our experiments, we observed that different models exhibit varying tendencies toward \emph{Unknown} responses. For instance, GPT-o4-mini is expected to have stronger reasoning capabilities than GPT-4o, yet both models achieve similar average accuracy rates. This occurs because the GPT-o4-mini model overthinks and produces too many \emph{Unknown} responses, thereby reducing its accuracy rate.

We additionally calculated the proportion of correct answers within valid responses, i.e., Precision. We found that GPT-o4-mini demonstrates significantly higher Precision score compared to other models, with a notable advantage in \fbeta as well. Detailed comparison can be found in Figure~\ref{fig:fbeta_comparison}.

\subsection{Does model really learn the logical paradigm?}

A persistent question is whether models merely memorize logical puzzles mechanically or truly understand their underlying logic.There is a logical consistency challenge~\cite{cheng2025empowering} in existing models, where models generally produce self-contradictory conclusions when faced with propositions sharing identical logical structures. However, \logictree naturally provides a convenient testing scenario for this challenge. During construction, multiple problems share the same logical tree structure, differing only in their natural language instantiations. Subsequently, we measured whether models could correctly solve all problems with the same logical structure, calculating Consistency Ratio to evaluate models' capacity to grasp the underlying principles.

As shown in Table~\ref{tab:consistency}, most models can understand simple deductive reasoning logic, but at reasoning depths of 7-8, even state-of-the-art models such as GPT-o3-mini, Qwen3-235B, deepseek-r1, and Claude3.7 demonstrate consistency rates approaching zero, revealing current models' insufficient capability for consistent extended thinking and complex combinatorial logic.

Additionally, we analyzed whether models explicitly utilized certain deductive reasoning rules in their responses. Figure~\ref{fig:word_frequency} presents word frequency statistics and sample responses from Qwen3-235B and GPT-o3-mini, showing that both models can identify implicit logical rules in the questions and apply these rules in their answers. However, as logical complexity increases, models gradually fail to recognize the underlying logic, with decreasing frequency of mentioning reasoning paradigms. Moreover, models exhibit varying capabilities in recognizing different logical paradigms. For instance, in deepseek-r1 responses, "\emph{Modus Tollens}" appears with the highest frequency, while "\emph{Disjunction Elimination}" occurs significantly less frequently. This discrepancy may result from the latter's more complex logic being harder to identify, or from insufficient exposure to this logical form during model pre-training.

\subsection{Does model's reasoning behaviour become more effective?}
\label{sec:val_log}
To isolate the effect of explanatory CoT on answer confidence, we test \textsc{Qwen2.5-7B-Instruct-1M} on $500$ randomly sampled GSM8K and $500$ LogicTree problems, generating for each prompt (i) a direct answer (No-CoT) and (ii) a step-by-step CoT that ends with the answer in a delimited span.We mark a CoT as \textbf{effective} if the model is \emph{incorrect}
in the No-CoT setting but \emph{correct} once the CoT is included.For each sample we compute  $\ell_{\text{CoT}} - \ell_{\text{NoCoT}}$. The mean log-probability gain of the ground-truth answer tokens $a_t^\star$ when conditioning on the CoT.
We further bucket the samples into four categories
\textbf{(WR)} wrong No-CoT / right CoT,
\textbf{(RR)} right No-CoT / right CoT,
\textbf{(WW)} wrong in both,
and \textbf{(RW)} right No-CoT / wrong CoT.
Table~\ref{tab:delta_transitions_gsm8k} and Table~\ref{tab:delta_transitions_logictree} report the statistics.

Across both benchmarks the \textbf{WR} group exhibits the largest
positive $\Delta\log p$ on average $+2.46$ nats for GSM8K and
$+1.8$ nats for LogicTree—confirming that
\emph{effective} CoTs substantially raise the model’s confidence in the correct answer.
By contrast, cases where the model is already correct (\textbf{RR})
yield only marginal gains, and \textbf{WW} often shows negative shifts,
suggesting that spurious reasoning can even erode confidence. In \textbf{(RW)} cases ,it's clearly that CoT tokens are negative.

We train the same base model with two popular base algorithms,
GRPO and DAPO on DRER framework. Picture ~\ref{fig:reasoning_quality} shows that reasoning quality reward increases sharply ,which proves that the CoT tokens are becoming positive for model reasoning.

Figure~\ref{fig:prediction_distribution_dapo} and figure~\ref{fig:prediction_distribution_base} plot the
prediction distribution  for a difficulty-3 problem from 100 samples.  Compared with the
DAPO 150 step baseline, the \drer-trained 150 step policy produces a markedly sharper peak
around the ground-truth answer, indicating that the learnt reasoning tokens help concentrate probability mass on the correct solution. Finally, Figure~\ref{fig:dapo_response} and figure~\ref{fig:grpo_response} show that
\drer keeps the average response length stable at fewer tokens,
saving tokens per problem relative to the baseline while
still achieving higher accuracy.  This validates \drer ability to
simultaneously improve reasoning quality and reduce inference cost.

\subsection{Does model acquire the generalization ability?}

Experiments in Table~\ref{tab:ood_performance} test benchmark sets that may require similar logical capabilities, demonstrating that models trained via the \drer framework to enhance logical reasoning abilities simultaneously exhibit certain generalization capabilities. On the ZebraLogic~\citep{lin2025zebralogic} and ProntoQA~\citep{propontoqa} logical datasets, models trained for 400 steps based on DAPO+\drer show modest improvements. ZebraLogic~\citep{} is constructed based on a grid logic puzzle, with the solution process equivalent to solving a SAT~\citep{sempolinski2009automatic} problem; while ProntoQA~\citep{propontoqa} consists of first-order deductive logic and hierarchical ontology, similar to the underlying principles of our \logictree, though its questions remain related to world knowledge. On OOD (out-of-distribution) datasets AIME24 and MMLU-redux~\citep{wang2024mmlu}, our models still demonstrate improvements over the base models, with pass@32 scores on AIME24 reaching the level of QwQ32B~\citep{dapo}.

\section{Conclusion and Future Work}
In this work, We present \drer, a token-level reward that ties each reasoning step to the model’s confidence, and introduce \logictree, a benchmark for formal deduction. On \logictree and other tasks, \drer yields more consistent and accurate chain-of-thoughts than existing rewards. The post-trained model also generalizes to unseen datasets, indicating that broader logical training further strengthens its reasoning ability.

\drer builds reasoning quality directly into the RL objective. With bounded rewards, each policy update provably raises the expected return, while a length-aware advantage keeps responses from growing erratic. This pairing steers the search toward chains that truly boost answer confidence without inflating cost. Because \drer needs no value model, it drops into existing pipelines and scales to larger or multimodal models. As future work, our aim is to extend the framework to further refining reasoning‑aligned training.

We release all code and the complete \logictree corpus to ensure transparency and reproducibility. Together, \drer and \logictree provide a lightweight, theoretically grounded basis for reasoning-aligned RL, enabling safer and more interpretable LLMs in logic-critical domains. Future work should extend this framework to richer logics and multimodal data.

\medskip

{
\small
\bibliographystyle{plain}
\bibliography{references}
}


\appendix

\section{Technical Appendices and Supplementary Material}
\begin{table}[ht]
  \centering

  \caption{An example of a logictree puzzle.}
  
  \begin{tcolorbox}[
      colframe=orange!70!black,    
      colback=orange!10!white,     
      sharp corners,             
      boxrule=0.5mm,             
      width=\textwidth,          
      title={An example of a logictree puzzle}
  ]
  \textbf{Paragraph}:\\
  On the condition that coral reefs need light to grow so only occur in shallow waters, it is definitely the case that in addition to this, olive oil is also ideal for frying and is the most stable fat when heated.If in addition to this, olive oil is also ideal for frying and is the most stable fat when heated, then if ribbons relate post : How to sew trims, then titanium dioxide and zinc oxide also functioned in this fashion.It is a fact that either the anus of this invertebrate is located on top of its body or coral reefs need light to grow so only occur in shallow waters.The statement that 'the anus of this invertebrate is located on top of its body' is incorrect. \\[1ex]

  \textbf{Question}:\\
  It is a common misconception that if ribbons relate post : How to sew trims, then titanium dioxide and zinc oxide also functioned in this fashion.\\[1ex]

  \textbf{Solution}:\\
  False
  \end{tcolorbox}

  \label{tab:logictree_puzzle}
\end{table}

\begin{table}[ht]
  \centering
   \caption{Model Response of logictree.}
  \begin{tcolorbox}[
      colframe=orange!70!black,   
      colback=orange!10!white,     
      sharp corners,            
      boxrule=0.5mm,             
      width=\textwidth,          
      title={Model Response of logictree} 
  ]
\textbf{Paragraph}:\\When the notion that 'if the statement that states the worms also eat the food scraps and worm bin bedding is false, then the statement 'emergent wetland vegetation is rooted in soil that is under the water for most of the time' can be considered false' is untrue is true, it follows that hydrangeas need minimal care in well-drained, fertile soil, and are shade lovers.One may reasonably assume that if the notion that 'if the statement that states the worms also eat the food scraps and worm bin bedding is false, then the statement 'emergent wetland vegetation is rooted in soil that is under the water for most of the time' can be considered false' is untrue, then those that suggest hydrangeas need minimal care in well-drained, fertile soil, and are shade lovers are wrong.Assuming sometimes, cancer cells left behind after treatment eventually form new tumors, we know that if either eggs from common ducks require about 28 days to hatch or the island was originally inhabited by Ligures Ilvates, who gave it the ancient name Ilva, then if the statement that states the worms also eat the food scraps and worm bin bedding is false, then emergent wetland vegetation is rooted in soil that is under the water for most of the time.The statement that states sometimes, cancer cells left behind after treatment eventually form new tumors is right.Whenever the claim that either eggs from common ducks require about 28 days to hatch or the island was originally inhabited by Ligures Ilvates, who gave it the ancient name Ilva is false is true, the following statement is also true: the bowl acts as a carb hole.Whenever the claim that either eggs from common ducks require about 28 days to hatch or the island was originally inhabited by Ligures Ilvates, who gave it the ancient name Ilva is false is true, the following statement is also true: the statement 'the bowl acts as a carb hole' can be considered false.\\
[1ex]\textbf{Question}:\\The statement 'the worms also eat the food scraps and worm bin bedding' can be considered true.\\It is a common misconception that if the statement that states the worms also eat the food scraps and worm bin bedding is false, then emergent wetland vegetation is rooted in soil that is under the water for most of the time.\\
[1ex]\textbf{Model Response}:\\1. Analyze the logical structure of the given statements.\\2. Identify the conditions and implications in each statement.\\3. Determine the truth value of the given statements and how they relate to each other.\\4. Check the consistency of the statements with the provided premises.\\5. Deduce the implications of the given statements on the conclusion.\\6. Evaluate the final conclusion based on the logical consistency and the given premises.\\\texttt{</think>}\\\texttt{<answer>[True, False]</answer>}\\
[1ex]\textbf{Solution}:\\True\\False
  \end{tcolorbox}
  \label{tab:logictree_puzzle2}
\end{table}

\begin{table}[ht]
  \centering
  \caption{Full Chain‑of‑Thought (CoT) Prompt Template Used for DRER Training and Evaluation.}
  \begin{tcolorbox}[
      colframe=teal!70!black,    
      colback=teal!10!white,     
      sharp corners,             
      boxrule=0.5mm,             
      width=\textwidth,          
      title={COT prompt}  
  ]
  \textbf{System Input}:\\
<|im\_start|>system

You are a helpful assistant. The assistant first thinks step by step about the reasoning
process in the mind and then provides the user with the answer.\newline
The reasoning process and answer are enclosed within \textless think\textgreater\ … \textless/think\textgreater\ and
\textless answer\textgreater\ … \textless/answer\textgreater\ tags, respectively, i.e.\newline
\textless think\textgreater\ Write the reasoning process for the given paragraph here \textless/think\textgreater\newline
\textless answer\textgreater\ Fill in the final answer list for \varnumq{} question(s) here: True, False or Unknown.
Like this: [True, False…] \textless/answer\textgreater

You must choose one of the following answers:\newline
-- TRUE: if the premises entail the statement\newline
-- FALSE: if the premises contradict the statement\newline
-- UNKNOWN: if you cannot determine the truth value of the statement from the premises

You will be given a paragraph of logical premises and a statement. Perform logical reasoning
\textbf{strictly based on the premises} using propositional logic.\newline
Assume all premises are true. Do not rely on prior world knowledge.

<|im\_end|>\\[1ex]

  \textbf{User Input}:\\
  <|im\_start|>user

Paragraph: \varparagraph{}

\varquestion{}

<|im\_end|>
<|im\_start|>assistant
<think>\\[1ex]

\textbf{Variable meanings}:\\
 \texttt{\{num\_q\}}: Number of questions in the current prompt.\\
 \texttt{\{paragraph\}}: The paragraph containing the logical premises.\\
 \texttt{\{current\_question\}}: The specific statement whose truth value is to be evaluated.\\
  \end{tcolorbox}

  \label{tab:COT_prompt}
\end{table}

\begin{table}[ht]
  \centering
  \caption{Full No‑CoT Prompt Template used for DRER training and evaluation.}
  \begin{tcolorbox}[
      colframe=teal!70!black,
      colback=teal!10!white,
      sharp corners,
      boxrule=0.5mm,
      width=\textwidth,
      title={No‑CoT Prompt}
  ]
  \small                      
  \textbf{System Input}:\\
\texttt{<|im\_start|>system}

\texttt{You are a helpful assistant. You answer questions by solely using logical reasoning.}\\
\texttt{You will be given a paragraph of logical premises and a statement. Perform logical reasoning \textbf{strictly based on the premises} using propositional logic.}\\
\texttt{Assume all premises are true. Do not rely on prior world knowledge.}\\[2pt]

\texttt{<answer> Fill in the final answer list for \varnumq{} question(s) here: True, False or Unknown. Like this: [True, False...] </answer>}\\
\texttt{You must choose one of the following answers:}\\
\texttt{\quad-- TRUE: if the premises entail the statement}\\
\texttt{\quad-- FALSE: if the premises contradict the statement}\\
\texttt{\quad-- UNKNOWN: if you cannot determine the truth value of the statement based on the premises}\\

\texttt{<|im\_end|>}\\[1ex]

\textbf{User Input}:\\
\texttt{<|im\_start|>user}

\texttt{Paragraph: \varparagraph}

\texttt{\varquestion}

\texttt{<|im\_end|>}\\
\texttt{<|im\_start|>assistant}\\
\texttt{<answer>} … \texttt{</answer>}\\[1ex]

\textbf{Variable meanings}:\\
\texttt{\{num\_q\}}: Number of questions in the current prompt.\\
\texttt{\{paragraph\}}: Paragraph containing the logical premises.\\
\texttt{\{current\_question\}}: Statement whose truth value is to be evaluated.\\
  \end{tcolorbox}

  \label{tab:NoCOT_prompt}
\end{table}

\begin{algorithm}[t]
\caption{\textbf{DRER}: \textbf{D}ynamic \textbf{R}easoning \textbf{E}fficiency \textbf{R}eward.}
\label{alg:drer}
\begin{algorithmic}[1]
  \Require
        Prompts $P=\{q_b\}_{b=1}^{B}$, ground-truth answers $Y^\star=\{a_b^\star\}_{b=1}^{B}$,\\
        policy $\pi_\theta$, rule reward $R_{\text{rule}}(\cdot)$, reasoning weight $\lambda_q$,\\
        bucket IDs $\{d_b\}_{b=1}^{B}$, bounds $\bigl(L^{(d)}_{\min},L^{(d)}_{\max}\bigr)$, temperature $\tau$
  \Ensure  Advantages $A\in\mathbb R^{B\times L}$

  \Statex\textbf{(1) Build trajectories}
  \State $C \gets \pi_\theta(P,\text{mode}=\textit{cot})$ \Comment{CoT trajectories}
  \For{$b=1$ \textbf{to} $B$}
      \State $t_n[b] \gets \textsc{NoCoTPrompt}(q_b) \,\Vert\,\textsc{FormatAnswer}(a_b^\star)$
      \State Replace answer span in $C[b]$ with $a_b^\star$ $\rightarrow\,t_c[b]$; record span $\mathcal A_b$
  \EndFor

  \Statex\textbf{(2) Reasoning-quality reward}
  \For{$b=1$ \textbf{to} $B$}
      \State $\ell_c=\frac{1}{|\mathcal A_b|}\!\sum_{t\in\mathcal A_b}\log p_\theta(a_{b,t}^\star\mid t_c[b])$
      \State $\ell_n=\frac{1}{|\mathcal A_b|}\!\sum_{t\in\mathcal A_b}\log p_\theta(a_{b,t}^\star\mid t_n[b])$
      \State $R_q[b]\gets\tanh(\ell_c-\ell_n)$
      \State $R_{\text{seq}}[b]\gets R_{\text{rule}}(C[b])+\lambda_q R_q[b]$
      \State Expand $R_{\text{seq}}[b]$ to token reward $r_{b,\cdot}$ on $C[b]$
  \EndFor

  \Statex\textbf{(3) Group-wise normalisation}
  \ForAll{prompt group $g$}
      \State $\mu_g \gets \mathrm{mean}(r_{m,\cdot}),\;
             \sigma_g \gets \mathrm{std}(r_{m,\cdot})\quad(m\!\in\! g)$
      \For{$m\in g$}  \Comment{raw advantage $\tilde A$}
          \State $\tilde A_{m,\cdot}\gets\dfrac{r_{m,\cdot}-\mu_g}{\sigma_g+\varepsilon}$
      \EndFor
  \EndFor

  \Statex\textbf{(4) Dynamic-length attenuation}
  \For{$b=1$ \textbf{to} $B$}
      \State $\ell_b\gets\textsc{Length}(C[b])$, \; $d\gets d_b$
      \State $g_b\gets\exp\!\Bigl(-\dfrac{\max\{0,\;L^{(d)}_{\min}-\ell_b,\;\ell_b-L^{(d)}_{\max}\}}{\tau}\Bigr)$
      \State $A_{b,\cdot}\gets g_b\cdot\tilde A_{b,\cdot}$
  \EndFor
  \State \Return $A$
\end{algorithmic}
\label{appendix:DRER}
\end{algorithm}

\section{\logictree: Formal Paradigms and Templates}
\label{appendix:logictree}

\subsection{Seven Deductive Paradigms in \textsc{\logictree}}
\label{appendix:paradigms}

\logictree centres on seven classic deductive paradigms that constitute the
atomic reasoning units of every sample.  Each paradigm is implemented as a
dedicated Python class (see \verb|logic.py|) whose constructor generates the
required premises and the logically entailed conclusion.  The table below
summarises their formal schemata together with bilingual surface examples.

\begin{table}[h]
  \centering
  \small
  \caption{Seven deductive paradigms that serve as the atomic reasoning units in \textsc{\logictree}.}
  \begin{tabular}{%
      @{}%
      p{0.22\linewidth}
      p{0.18\linewidth}
      >{\raggedright\arraybackslash}p{0.58\linewidth}
      @{}%
    }
    \toprule
    \textbf{Paradigm} & \textbf{Formal Schema} & \textbf{Surface Realisation} \\ \midrule
    Modus Ponens &
    $(p\!\rightarrow\!q)\land p \;\Rightarrow\; q$ &
    If Alice studies, she will pass. Alice studies. Therefore, she will pass. \\[1ex]\midrule
    Modus Tollens &
    $(p\!\rightarrow\!q)\land\neg q \;\Rightarrow\; \neg p$ &
    If it rains, the road is wet. The road is not wet. Thus, it did not rain. \\[1ex]\midrule
    Hypothetical Syllogism &
    $(p\!\rightarrow\!q)\land(q\!\rightarrow\!r) \;\Rightarrow\; (p\!\rightarrow\!r)$ &
    If A wins, B celebrates. If B celebrates, C is happy. Hence, if A wins then C is happy.\\[1ex]\midrule
    Disjunctive Syllogism &
    $(p\lor q)\land\neg p \;\Rightarrow\; q$ &
    Either today is Monday or Tuesday. Today is not Monday. Therefore, today is Tuesday. \\[1ex]\midrule
    Reductio ad Absurdum &
    $(p\!\rightarrow\!q)\land(p\!\rightarrow\!\neg q) \;\Rightarrow\; \neg p$ &
    Assume the number is both even and odd. This leads to a contradiction. Thus, the number is not both even and odd. \\[1ex]\midrule
    Constructive Dilemma &
    $(p\!\rightarrow\!q)\land(r\!\rightarrow\!s)\land(p\lor r) \;\Rightarrow\; (q\lor s)$ &
    If it rains, we stay in; if it is sunny, we picnic. Either it rains or it is sunny. Hence, we either stay in or picnic. \\[1ex]\midrule
    Disjunction Elimination &
    $(p\lor q)\land(p\!\rightarrow\!s)\land(q\!\rightarrow\!s) \;\Rightarrow\; s$ &
    Either I study or I work. If I study, I will learn. If I work, I will learn. Thus, I will learn. \\
    \bottomrule
  \end{tabular}
  
  \label{tab:paradigms}
\end{table}

\paragraph{Instantiation pipeline.}
For every sampled instance, the generator (i) uniformly selects one of the
seven paradigms, (ii) instantiates the schematic variables with randomly
sampled entities, and (iii) lexicalises the symbolic premises and conclusion
using template pools containing 10–12 paraphrases per construct.  Optional
nesting of paradigms allows multi-step reasoning chains while preserving
formal validity via internal consistency checks.

\subsection{Primitive and Compound Propositions}
\label{appendix:compounds}

\logictree expresses every deductive instance in terms of one \emph{primitive
statement} and four \emph{compound connectives}.  The primitive
\verb|Statement| captures an atomic fact—e.g.\ “Alice studies.”—
while the four connectives build larger formulas: \textit{negation},
\textit{conjunction}, \textit{implication}, and \textit{inclusive
disjunction}.  Each connective is implemented as a dedicated class whose
method \verb|.nl()| randomly selects a surface template from
\texttt{expressions.json}.  \autoref{tab:compound-types} summarises the five
constructs, their formal notation, and representative English renderings.

\begin{table}[h]
  \centering
  \small
  \caption{Primitive and compound proposition types used in \textsc{LogicTree}.}
  \begin{tabular}{c c c}
    \toprule
    \textbf{Construct} & \textbf{Logical Form} & \textbf{Example Surface Realisation (EN)} \\ \hline
    Statement (atomic) &
    $p$ &
    \textit{Alice studies.} \\ \hline
    Negation &
    $\lnot p$ &
    \textit{It is \textbf{not} true that Alice studies.} \\ \hline
    Conjunction &
    $P \wedge q$ &
    \textit{Alice studies \textbf{and} Bob plays chess.} \\ \hline
    Implication (Conditional) &
    $P \rightarrow q$ &
    \textit{\textbf{If} it rains, \textbf{then} the road becomes wet.} \\ \hline
    Inclusive Disjunction &
    $P \lor q$ &
    \textit{\textbf{Either} today is Monday \textbf{or} Tuesday.} \\ \bottomrule
  \end{tabular}
  \label{tab:compound-types}
\end{table}


\paragraph{Surface realisation.}
When generating a sample, the pipeline first creates
\verb|Statement| objects for the chosen entities, then composes them with the
connectives above.  For example, calling
\verb|Negation(S).nl()| yields a randomly chosen negated template such as
\emph{“The claim that {S} is false.”}; calling
\verb|Conditional(P,Q).nl()| may return \emph{“Provided that {P}, we know
that {Q}.”}.  This template sampling, combined with optional adverb or
negator insertion, gives \logictree a high level of lexical diversity while
preserving formal truth values.

\subsection{Training setting}
\label{appendix:training_setting}
Table ~\ref{tab:training_params} records important training parameters.
Experiments are conducted on 4×H20 (80G) GPUs with CUDA 12.0, PyTorch 2.6.0, transformers 4.47.1. The Main Experiment phase (DAPO+\drer) trains for 400 training steps and takes approximately 50 hours. Training is carried out with a learning rate of \(3 \times 10^{-7}\), a maximum response length of \(4096\) tokens, the batch size is \(16\) and \(16\) responses per prompt. For GRPO, the KL divergence coefficient is set to \(0.001\). In the \drer framework, we set \(\lambda_q = 1\) and \(\tau = 8\). 
\begin{table}[ht]
\centering
\caption{Important Training Parameters.}
\begin{tabular}{ccccc}
\toprule
\textbf{Algorithm} & \textbf{Train Batch Size} & \textbf{Rollout N} & \textbf{KL Coef} & \textbf{Max Response Len} \\
\midrule
GRPO  & 16 & 16 & 0.001 & 4096 \\
DAPO  & 16 & 16 & --     & 4096 \\
\bottomrule
\end{tabular}
\label{tab:training_params}
\end{table}

\section{Related dataset}
Logical reasoning datasets can broadly be categorized into three types. The first type focuses on deductive reasoning. The second type is based on grid-based logic puzzles. The third category comprises datasets based on multi-hop or strategic question answering. These datasets assess language models’ logical capabilities from various perspectives, including formal logic, multi-step planning, structural induction, and strategy analysis. In addition, there are general-purpose reasoning datasets that are also frequently used to evaluate LLMs' logical reasoning abilities.

\subsection{Deductive Reasoning}

ConTRoL~\citep{Control}, consisting of 8,325 pairs of expert-designed datasets, is a challenging segment-level NLI dataset to evaluate model’s contextual reasoning capacity from police recruitment tests. RuleTaker~\citep{ruletaker} is a benchmark dataset designed to test whether language models can logically reason about natural language rules and facts by determining whether the conclusions follow, do not follow, or are uncertain. LogiQA~\citep{logiqa} is a benchmark of 8,678 civil service exam questions designed to evaluate models’ reading comprehension and deductive reasoning across five logical types by requiring conclusion drawing from textual premises. LogiQA2.0~\citep{logiqa2.0} is the enchanced version of LogiQA~citep{logiqa}, featuring improved translations, expert-verified annotations, and new NLI tasks, designed to evaluate logical reasoning and reading comprehension in MRC and NLI formats. FOLIO~\cite{folio} is an maually annotated dataset containing 1,430 logically complex natural language reasoning examples with first-order logic (FOL) annotations, designed to evaluate and benchmark the deductive reasoning and NL-FOL translation capabilities of Large Language models. PrOntoQA~\citep{propontoqa} is a benchmark proposed in 2022 to evaluate LLMs’ reasoning by generating question-answer pairs from first-order logic, revealing their struggles with multi-step proof planning despite valid individual steps. Compared with PrOntoQA~\citep{propontoqa}, PrOntoQA-OOD~\citep{prompontoqaood} is designed to evaluate the general deductive reasoning abilities of LLMs by testing their ability to generalize to more complex, compositional proofs, particularly those that are out-of-distribution (OOD). JustLogic~\citep{justlogic} a generated deductive reasoning benchmark designed to evaluate LLMS, featuring high complexity, being independent of prior knowledge, and conducting in-depth error analysis in terms of reasoning depth and argumentative form.

However, the existing logical reasoning datasets still have some limitations. Most datasets have fixed or limited reasoning depth and breadth, which limits their ability to conduct a comprehensive evaluation of complex multi-step reasoning models. Many datasets entwine semantic information with logic, which may lead the model to rely on semantic cues rather than pure logical reasoning. 

Furthermore, the majority focus only on final answer correctness, lacking assessment of the intermediate reasoning process and overall explanation quality.

In contrast, the LogicTree dataset we proposed has significant advantages: it is programmed and dynamically constructed, allowing for flexible control over the depth, breadth, and difficulty of inference; It separates semantics from logic to precisely evaluate pure deductive reasoning;  It introduces a new logical consistency metric across multiple logical equivalence problems to measure the model's grasp of the underlying logical structure.

\subsection{Grid-Based Logic Puzzles}

BoardgameQA\citep{boardgameqa} is a dataset designed to evaluate the reasoning ability of language models when dealing with contradictory information.GridPuzzle\citep{gridpuzzle} is a dataset of grid-based logic puzzles designed to evaluate LLMs’ structured, multi-step reasoning abilities through both final answers and detailed reasoning chains.The Knights and Knaves\citep{xie2025logicrlunleashingllmreasoning} dataset is an reasoning dataset designed to test logical deduction, where characters are either knights (truth-tellers) or knaves (liars), featuring controlled difficulty levels, procedural generation, and verifibility. 

Existing datasets, such as GridPuzzle~\citep{tyagi2024step}, Knights and Knaves (KK)~\citep{xie24memorization} provide valuable reasoning benchmarks, but they all have limitations. For example, KK~\citep{xie24memorization} entangles logical reasoning with semantic cues, taking the risk of rapid learning through keyword associations. Some logic puzzle focuses on the final answer without verifying the intermediate steps, allowing the model to guess without sufficient reasoning.

On the contrary, \logictree evaluates the final and intermediate steps and executes the complete reasoning chain. It also introduces a logical consistency rate among variants of the same logical form and uses semantic-logical unentanglement to ensure that the model relies on reasoning rather than superficial clues.

\subsection{Multi-hop or Strategic Question Answering}
HotpotQA~\citep{yang2018hotpotqa} is a multi-hop question-answering dataset that requires reasoning across multiple documents and provides supporting facts to enhance the interpretability of the QA system.StrategyQA~\citep{strategtqa} is a benchmark dataset designed to evaluate implicit multi-step reasoning in LLMs across 15 domains and 13 strategies. SPAG~\citep{cheng2024self} is self-laying based adversarial language game dataset designed to enhance and evaluate the reasoning ability through a game involving indirect communication and strategic reasoning about hidden target words. LOGICGAME~\citep{gui2024logicgame} is a benchmark designed to evaluate LLMs’ ability to understand, execute, and plan based on predefined rules through diverse, verifiable game scenarios requiring multi-step logical reasoning.AutoLogi~\citep{zhu2025autologi} is benchmark test for open-ended logic puzzles with controllable difficulty and program-based verification, designed to evaluate the reasoning ability of LLM.

Compared with datasets such as HotpotQA~\citep{yang2018hotpotqa}, StrategyQA~\citep{strategtqa}, they emphasize various forms of multi-step or strategic reasoning across natural language problems, but there are still obvious limitations: The reasoning strategies in existing datasets are often broad and empirical rather than based on formal logical deduction frameworks (for example, StrategyQA~\citep{strategtqa} relies on heuristic and empirical categories). Many datasets focus on language pattern matching or cross-document evidence aggregation rather than verifying the true formal reasoning process (for example, HotpotQA~\citep{yang2018hotpotqa}).\logictree, on the other hand, strictly adheres to classical mathematical logic, adopting clear and well-defined deduction rules, and does not rely on common sense knowledge, providing a purest logical reasoning environment.

\subsection{General purpose dataset}

MMLU-Pro\citep{wang2024mmlu} is an advanced benchmark of 12,000 expert-reviewed, 10-option questions across 14 disciplines, designed to better evaluate LLM performance with greater difficulty and reduced noise than the original MMLU~\citep{hendrycks2021measuring}. However, it primarily evaluates broad knowledge and reasoning abilities rather than focusing on strong formal logical reasoning. Thus, it is not specifically designed to test models’ capabilities in complex multi-step logical dedu

\section{Prompt Templates}
\label{appendix:prompts}

Tables~\ref{tab:COT_prompt} and~\ref{tab:NoCOT_prompt} list the exact
prompts used in our experiments: a \emph{Chain-of-Thought (CoT)} version
that elicits step-by-step reasoning, and a \emph{No-CoT} variant that asks
for the final answer only.  Curly-braced placeholders are replaced at
runtime (\texttt{\{paragraph\}}, \texttt{\{current\_question\}},
\texttt{\{num\_q\}}).  The two prompts share identical task instructions,
so performance differences isolate the effect of showing or hiding the
reasoning chain.

\section{Training details}
\label{appendix:Training Details}

\begin{table}[h]
  \centering
  \caption{Average $\ell_{\text{CoT}} - \ell_{\text{NoCoT}}$ by answer transition in GSM8K.}
  \begin{tabular}{c|cc}
    \toprule
    Original \(\downarrow\) / With CoT \(\rightarrow\) & Wrong (W) & Correct (R) \\
    \midrule
    Wrong (W) &  -4.32 &  2.46 \\
    Correct (R) &  -5.00 &  -0.47 \\
    \bottomrule
  \end{tabular}
  \label{tab:delta_transitions_gsm8k}
\end{table}

\begin{table}[h]
  \centering
  \caption{Average $\ell_{\text{CoT}} - \ell_{\text{NoCoT}}$ by answer transition in LogicTree.}
  \begin{tabular}{c|cc}
    \toprule
    Original \(\downarrow\) / With CoT \(\rightarrow\) & Wrong (W) & Correct (R) \\
    \midrule
    Wrong (W) &  -1.13 &  1.81 \\
    Correct (R) &  -3.79 &  -4.76 \\
    \bottomrule
  \end{tabular}
  \label{tab:delta_transitions_logictree}
\end{table}

\begin{figure}[h]
  \includegraphics[width=\linewidth]{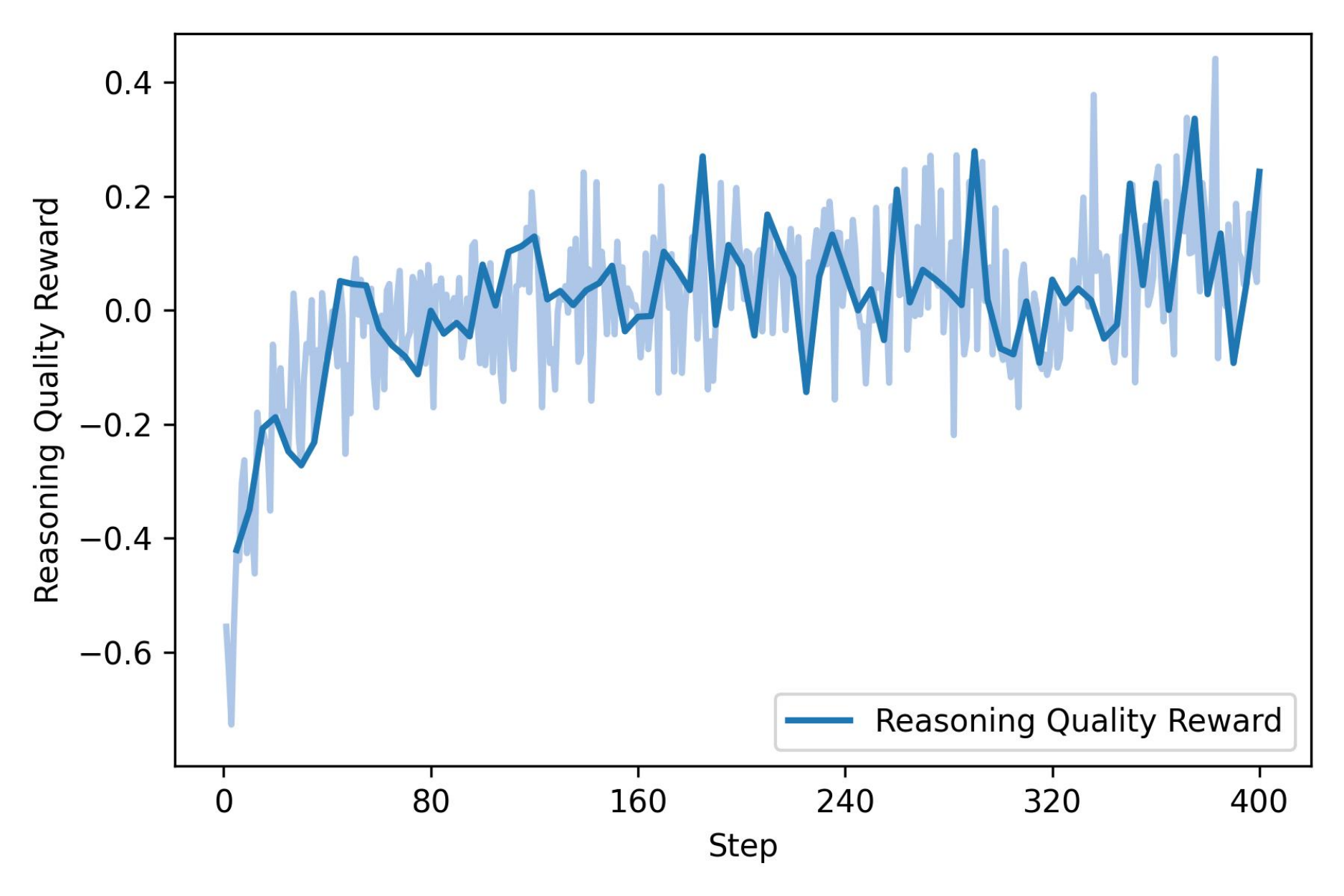}
  \caption{Reasoning quality reward on the LogicTree during post-training with \drer.}
  \label{fig:reasoning_quality}
\end{figure}

\begin{figure}[h]
  \includegraphics[width=\linewidth]{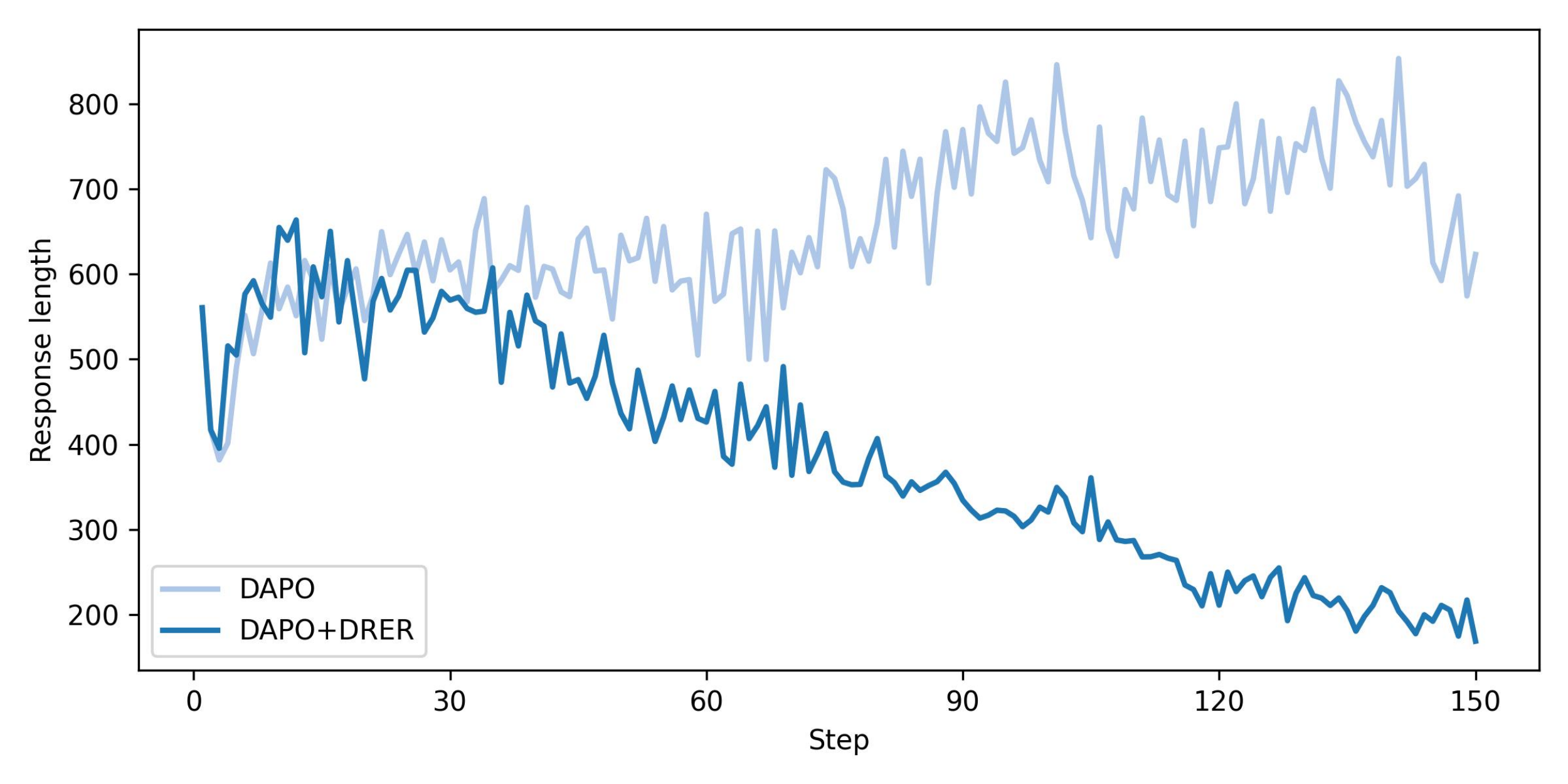}
  \caption{Comparison of response lengths over training steps between DAPO and DAPO+DRER. The integration of DRER leads to a reduction in response length, indicating enhanced efficiency with concise output.}
  \label{fig:dapo_response}
\end{figure}

\begin{figure}[h]
  \includegraphics[width=\linewidth]{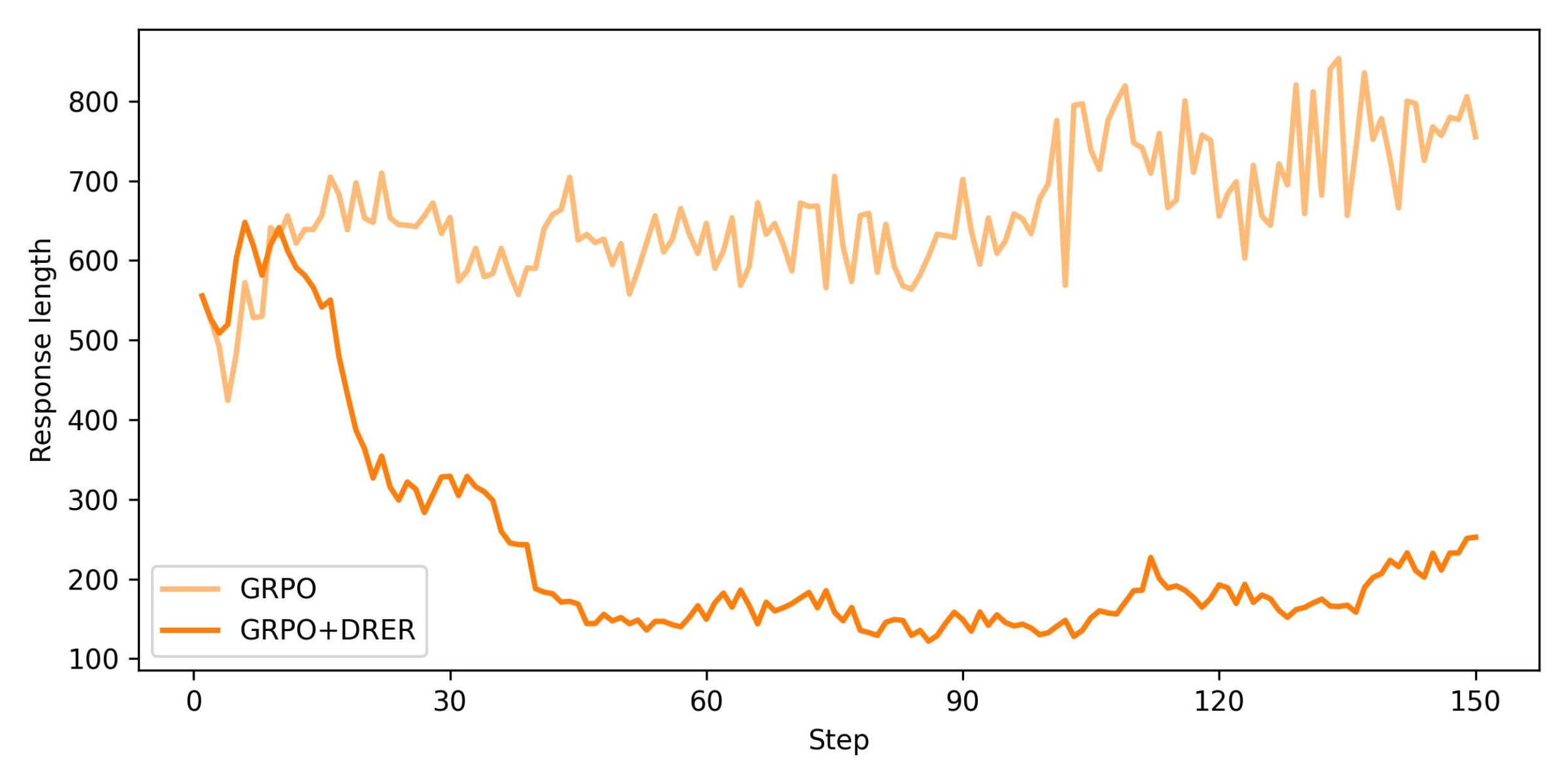}
  \caption{Comparison of response lengths over training steps between GRPO and GRPO+DRER. The integration of DRER leads to a reduction in response length, indicating enhanced efficiency with concise output.}
  \label{fig:grpo_response}
\end{figure}

\begin{figure}[h]
  \includegraphics[width=\linewidth]{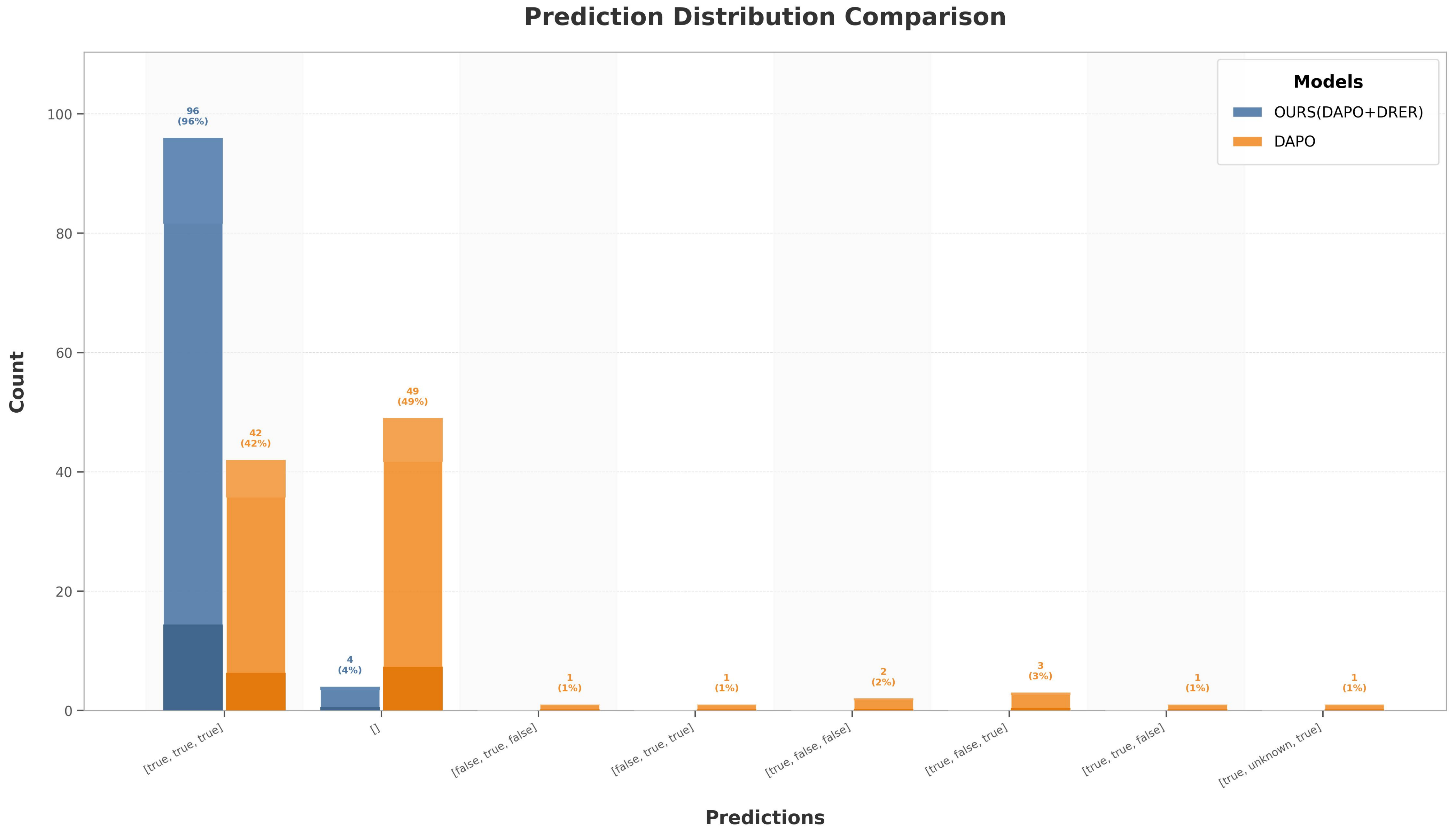}
  \caption{Prediction distribution comparison between DAPO and DAPO+DRER under Chain-of-Thought (CoT) prompting. The DAPO+DRER model demonstrates significantly higher confidence in correct answers, as shown by a strong concentration of predictions on the fully correct label set ([true, true, true]). In contrast, the baseline DAPO model produces more scattered outputs, indicating lower certainty. This highlights the effectiveness of DRER in combination with CoT reasoning for improving answer consistency and correctness.}
  \label{fig:prediction_distribution_dapo}
\end{figure}

\begin{figure}[h]
  \includegraphics[width=\linewidth]{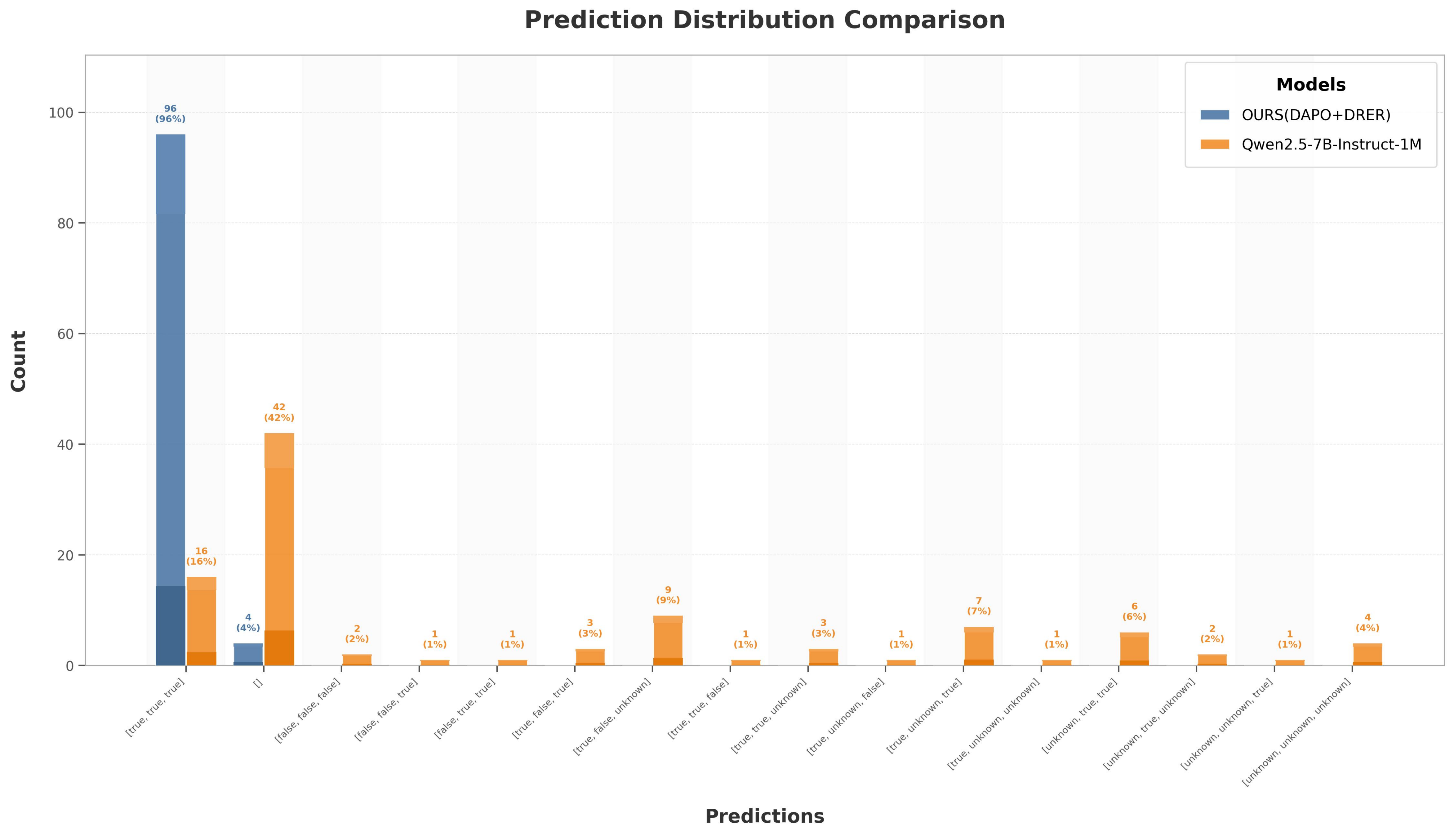}
  \caption{Prediction distribution comparison between our model (DAPO+DRER) and Qwen2.5-7B-Instruct-1M under Chain-of-Thought (CoT) prompting. The DAPO+DRER model produces highly concentrated predictions on the fully correct label ([true, true, true]), indicating strong confidence and consistency. In contrast, Qwen2.5-7B-Instruct-1M predictions are widely dispersed across incorrect and partially correct categories, reflecting lower answer certainty. This highlights the effectiveness of DRER combined with CoT in guiding the model toward accurate and confident output.}
  \label{fig:prediction_distribution_base}
\end{figure}

\section{Supplementary Evaluation}
\label{appendix:Supplementary Evaluation}

\begin{figure}[h]
  \includegraphics[width=\linewidth]{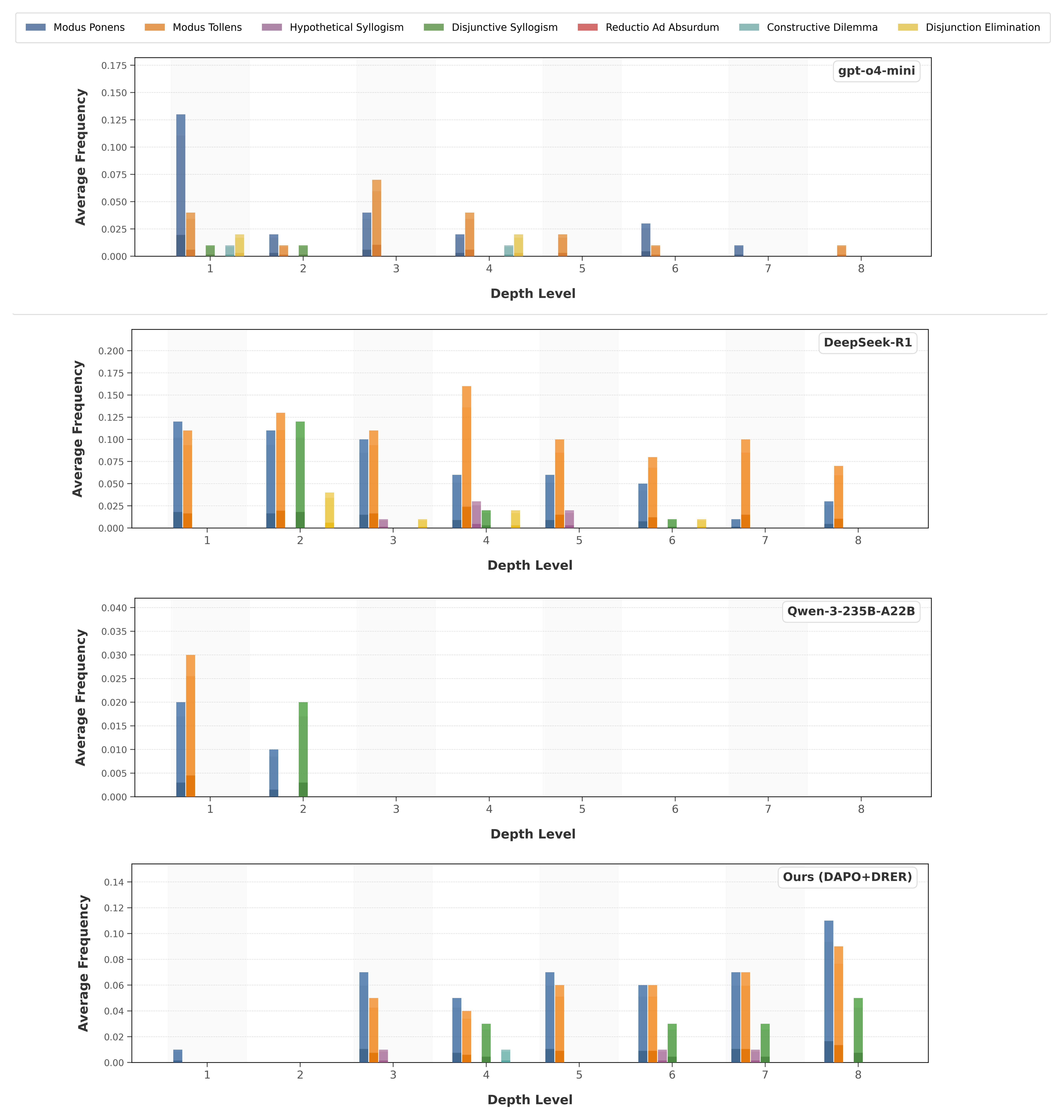}
  \caption{Word frequencies of seven deductive reasoning terms explicitly mentioned in LLMs response \drer.}
  \label{fig:word_frequency}
\end{figure}

\begin{figure}[h]
  \includegraphics[width=\linewidth]{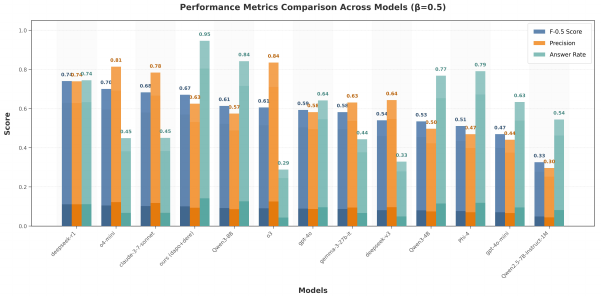}
  \caption{\fbeta, Answer Rate and Precision metrics Comparison across various models.}
  \label{fig:fbeta_comparison}
\end{figure}

Table~\ref{tab:consistency} presents the entire evaluation data of Consistency Ratio. Figure~\ref{fig:word_frequency} shows the distribution of those deduction logical key words in LLMs response. Figure~\ref{fig:fbeta_comparison} plot the complete evaluation data of \fbeta, which provides a balanced metric to compare the comprehensive performance across those LLMs. Tables~\ref{tab:ood_performance} records the evaluation results on other ood benchmarks, including AIME24, MMLU-redux~\citep{wang2024mmlu}, ZebraLogic~\citep{lin2025zebralogic} and ProntoQA~\citep{propontoqa}. 


\begin{table}[t]
  \centering
  \small
  \caption{Comparison of Consistency Ratio on \logictree across various logical depth.}
  \begin{tabular}{@{}l *{9}{c} @{}}
    \toprule
    Model & 1 & 2 & 3 & 4 & 5 & 6 & 7 & 8 & Avg. \\
    \midrule
    Qwen3-235B-A22B & \textbf{0.90} & 0.65 & 0.30 & \textbf{0.50} & 0.15 & 0.00 & \textbf{0.05} & 0.00 & 0.32 \\
    Deepseek-R1 & 0.70 & 0.55 & 0.20 & 0.15 & 0.10 & 0.00 & \textbf{0.05} & 0.00 & 0.22 \\
    Claude-3.7-Sonnet & 0.65 & 0.35 & 0.00 & 0.00 & 0.00 & 0.00 & 0.00 & 0.00 & 0.12 \\
    Qwen3-8B & 0.65 & 0.70 & 0.05 & 0.05 & 0.05 & 0.00 & 0.00 & 0.00 & 0.19 \\
    GPT-o4-mini & 0.50 & 0.35 & 0.00 & 0.05 & 0.00 & 0.00 & 0.00 & 0.00 & 0.11 \\
    GPT-o3-mini & 0.45 & 0.30 & 0.00 & 0.00 & 0.00 & 0.00 & 0.00 & 0.00 & 0.09 \\
    Qwen3-4B & 0.40 & 0.30 & 0.05 & 0.05 & 0.00 & 0.00 & 0.00 & 0.00 & 0.10 \\
    
    \midrule
    Gemini-2.5-Flash-Preview & 0.75 & 0.50 & 0.15 & 0.05 & 0.00 & 0.00 & 0.00 & 0.00 & 0.18 \\
    GPT-4o & 0.40 & 0.35 & 0.00 & 0.05 & 0.00 & 0.00 & 0.00 & 0.00 & 0.10 \\
    Phi-4-14 & 0.35 & 0.35 & 0.05 & 0.05 & 0.00 & 0.00 & 0.00 & 0.00 & 0.10 \\
    Gemma-3-27B & 0.25 & 0.20 & 0.00 & 0.00 & 0.00 & 0.00 & 0.00 & 0.00 & 0.06 \\
    Deepseek-v3 & 0.15 & 0.00 & 0.00 & 0.00 & 0.00 & 0.00 & 0.00 & 0.00 & 0.02 \\
    GPT-4o-mini & 0.10 & 0.10 & 0.00 & 0.00 & 0.00 & 0.00 & 0.00 & 0.00 & 0.03 \\
    \midrule

    Qwen2.5-7B-Instruct-1M & 0.05 & 0.00 & 0.00 & 0.00 & 0.00 & 0.00 & 0.00 & 0.00 & 0.01 \\
    \rowcolor{cyan!20}
    \textbf{Ours (DAPO+DRER)} & 0.70 & \textbf{0.70} & \textbf{0.60} & 0.35 & \textbf{0.50} & \textbf{0.35} & 0.00 & \textbf{0.10} & \textbf{0.41}\bbonus{0.40} \\
    \bottomrule
  \end{tabular}
  \label{tab:consistency}
\end{table}

\begin{table*}[h]
  \centering
  \caption{Performance of Qwen2.5 baseline and our model on various benchmarks.}
  \label{tab:ood_performance}
  \resizebox{0.95\textwidth}{!}{
  \begin{tabular}{lcccc}
    \toprule
    \textbf{Model} & \textbf{AIME24 (avg@32)} & \textbf{MMLU-redux~\citep{wang2024mmlu}} & \textbf{ZebraLogic~\citep{lin2025zebralogic} (Cell Acc)} & \textbf{ProntoQA~\citep{propontoqa}} \\
    \midrule
    Qwen2.5-Instruct-1M & 12.8 & 71.7 & 30.9 & 41.0 \\
    Ours (DAPO+\drer)     & 16.5 & 73.3 & 33.4 & 45.0 \\
    \bottomrule
  \end{tabular}
  }
\end{table*}

\begin{table}[!t]
\centering
\caption{Details of the organization and model source (model version for proprietary models, and Huggingface model name for open-source models) for the LLMs evaluated in \logictree.}
\resizebox{0.95\textwidth}{!}{%
\renewcommand{\arraystretch}{1.1}
\begin{tabular}{llllp{10cm}}
\toprule
Model & Organization & Size & Notes & Source \\
\midrule
DeepSeek-R1 & DeepSeek & 671B & MoE &  \texttt{deepseek-ai/DeepSeek-R1} \\
DeepSeek-V3 & DeepSeek & 671B & MoE &  \texttt{deepseek-ai/DeepSeek-V3} \\

Claude 3.7 Sonnet & Anthropic & -- & & \texttt{claude-3-7-sonnet-20250219} \\

Gemini 2.0 Flash Thinking Preview & Google & -- & & \texttt{gemini-2.5-flash-preview-04-17} \\
Gemma-3-27B & Google & 27B & & \texttt{google/gemma-3-27b-it} \\

Qwen3-235B-A22B & Alibaba & 235B & MoE & \texttt{qwen3-235b-a22b} \\
Qwen3-30B-A3B & Alibaba & 30B & MoE & \texttt{qwen3-30b-a3b} \\
Qwe3-8B & Alibaba & -- &  & \texttt{qwen3-8b} \\
Qwen3-4B & Alibaba & -- &  & \texttt{qwen3-4b} \\
Qwen2.5-7B-Instruct-1M & Alibaba & -- & MoE & \texttt{qwen2.5-7b-instruct-1m} \\

Phi-4-14B & Microsoft & 14B & & \texttt{microsoft/phi-4} \\

GPT-o4-mini & OpenAI & -- & & \texttt{o4-mini-2025-04-16} \\
GPT-o3 & OpenAI & -- & & \texttt{o3-mini-2025-01-31} \\
GPT-4o-mini & OpenAI & -- & & \texttt{gpt-4o-mini-2024-07-18} \\
GPT-4o & OpenAI & -- & & \texttt{gpt-4o-2024-11-20} \\
\bottomrule
\end{tabular}
}
\label{tab:model-detail}
\end{table}

\section{Limitations}

Despite the empirical gains achieved by DRER and LogicTree, several
limitations remain:

\begin{itemize}[leftmargin=15pt, itemsep=0pt, topsep=0pt, partopsep=0pt, parsep=0.5pt]
  \item \textbf{Logic coverage.} \logictree is limited to the deductive reasoning paradigm, while more diverse forms such as analogical reasoning, inductive reasoning, or traceable reasoning have not yet been evaluated.
  \item \textbf{Model scale and cost.} All experiments use Qwen-2.5-7B-Instruct-1M as backbone. The memory and latency overhead of token-level rewards on 70 B-scale or MoE models is unknown and may be prohibitive.
  \item \textbf{Evaluation bias.} Training and evaluation rely on an automatic logic verifier and confidence scores; no human preference or chain-quality annotation is included, which may overlook subjective aspects of reasoning quality.
  \item \textbf{Synthetic corpus and social bias.}  \logictree sentences are synthetically generated; potential social biases or misuse risks in real-world deployments have not been systematically analysed.
\end{itemize}

In future work we plan to extend DRER to higher-order logic, explore
low-cost reward approximations, and incorporate human evaluation and bias
auditing to mitigate these limitations.

\section{Broader Impact}

Our work aims to align large language models with formal logical
principles, potentially improving the reliability and interpretability of
machine reasoning.  By releasing the \textsc{LogicTree} dataset and DRER
code under a permissive licence, we enable researchers and practitioners to
build verifiable agents for education, scientific discovery, and safety-
critical auditing, where transparent deductive chains are preferable to
opaque heuristics.

\subsection{Positive societal outcomes.}
A reasoning-aligned model can serve as a didactic tutor in introductory
logic courses, assist engineers in detecting faulty assumptions in
software specifications, and support legal or medical professionals by
highlighting which premises lead to a conclusion rather than merely
producing an answer.  The synthetic nature of \textsc{LogicTree} limits
exposure to personal data and reduces the risk of privacy leaks.

\subsection{Potential risks.}
More persuasive and logically consistent outputs could be weaponised for
misinformation or overly authoritative automation.  Over-reliance on
synthetic benchmarks might also hide biases that appear in real-world
discourse.  Furthermore, token-level reward signals expose fine-grained
model behaviour, which could be exploited to reverse-engineer proprietary
system prompts.

\subsection{Mitigations.}
We distribute our resources with an explicit no-malicious-use clause,
encourage downstream users to apply bias and misinformation audits, and
recommend human oversight for high-stakes deployment.  Future work will
extend DRER to real-world corpora and incorporate human preference
feedback, allowing broader yet safer adoption of reasoning-aligned
reinforcement learning.

\end{CJK}
\end{document}